\documentclass[10pt,journal,compsoc]{IEEEtran}

\ifCLASSOPTIONcompsoc
  \usepackage[nocompress]{cite}
\else
  \usepackage{cite}
\fi

\usepackage{chngcntr}
\usepackage{multirow}
\usepackage{graphicx}
\usepackage{comment}
\usepackage{amsmath,amssymb} 
\usepackage{color}
\usepackage[dvipsnames]{xcolor}
\usepackage[american]{babel}
\usepackage[pagebackref=true,breaklinks=true,letterpaper=true,colorlinks,bookmarks=false]{hyperref}
\usepackage[utf8]{inputenc}
\usepackage{babel}
\usepackage[switch]{lineno}
\usepackage[font=small,labelfont=bf]{caption}
\usepackage{floatrow}
\usepackage{endnotes}
\usepackage{footnote}
\usepackage[]{footmisc}
\newcommand*{\Scale}[2][4]{\scalebox{#1}{$#2$}}%

\newcommand{\etal}{\textit{et al.}} 
\newcommand{\ie}{\textit{i.e.}}
\newcommand{\eg}{\textit{e.g.}}
\newcommand{\quotecap}[1]{``{\em #1}''}

\usepackage{lipsum}
\usepackage{graphicx}

\usepackage{mychangebar}
\newcommand{\red}[1]{\textcolor{black}{#1}}

\begin{document}


\title{On Distinctive Image Captioning via Comparing and Reweighting}

\author{Jiuniu~Wang$^{\star}$,
        Wenjia~Xu$^{\star}$,
        Qingzhong~Wang,
        and~Antoni~B.~Chan,~\IEEEmembership{Senior~Member,~IEEE}
\IEEEcompsocitemizethanks{\IEEEcompsocthanksitem J.~Wang, Q.~Wang, and A.~B.~Chan are with the Department Computer Science, City University of Hong Kong, Hong Kong.~Q.~Wang is also with Baidu Research, Baidu Inc., Beijing, China. \protect\\
E-mail: jiuniwang2-c@my.cityu.edu.hk, wangqingzhong@baidu.com, abchan@cityu.edu.hk
\IEEEcompsocthanksitem J.~Wang and W.~Xu are with Aerospace Information Research Institute, Chinese Academy of Sciences, and also University of Chinese Academy of Sciences, Beijing, China. W.~Xu is also with State Key Laboratory of Networking and Switching Technology, Beijing University of
Posts and Telecommunications, Beijing, China.\protect\\
E-mail: xuwenjia16@mails.ucas.ac.cn

\IEEEcompsocthanksitem $^{\star}$ indicates equal contribution.
}

\thanks{Manuscript received xxx; revised xx
	.}}

\markboth{IEEE TRANSACTIONS ON PATTERN ANALYSIS AND MACHINE INTELLIGENCE,~Vol.~xx, No.~xx, xx~xx}
{WANG \MakeLowercase{\textit{et al.}}: Distinctive image captioning}

\IEEEtitleabstractindextext{%
\begin{abstract}
Recent image captioning models are achieving impressive results based on popular metrics, \ie, BLEU, CIDEr, and SPICE. However, focusing on the most popular metrics that only consider the overlap between the generated captions and human annotation could result in using common words and phrases, which lacks distinctiveness, \ie, many similar images have the same caption. In this paper, we aim to improve the distinctiveness of image captions via comparing and reweighting with a set of similar images. First, we propose a distinctiveness metric---between-set CIDEr (CIDErBtw) to evaluate the distinctiveness of a caption with respect to those of similar images. Our metric reveals that the human annotations of each image in the MSCOCO dataset are not equivalent based on distinctiveness;  however, previous works normally treat the human annotations equally during training, which could be a reason for generating less distinctive captions. In contrast, we reweight each ground-truth caption according to its distinctiveness during training. We further integrate a long-tailed weight strategy to highlight the rare words that contain more information, and captions from the similar image set are sampled as negative examples to encourage the generated sentence to be unique. Finally, extensive experiments are conducted, showing that our proposed approach significantly improves both distinctiveness (as measured by CIDErBtw and retrieval metrics) and accuracy (\eg, as measured by CIDEr) for a wide variety of image captioning baselines. These results are further confirmed through a user study.
\end{abstract}

}

\maketitle

\IEEEdisplaynontitleabstractindextext

\IEEEpeerreviewmaketitle

\section{Introduction}
\label{sec:introduction}

\begin{figure*}
	\centering
	\includegraphics[height=6cm]{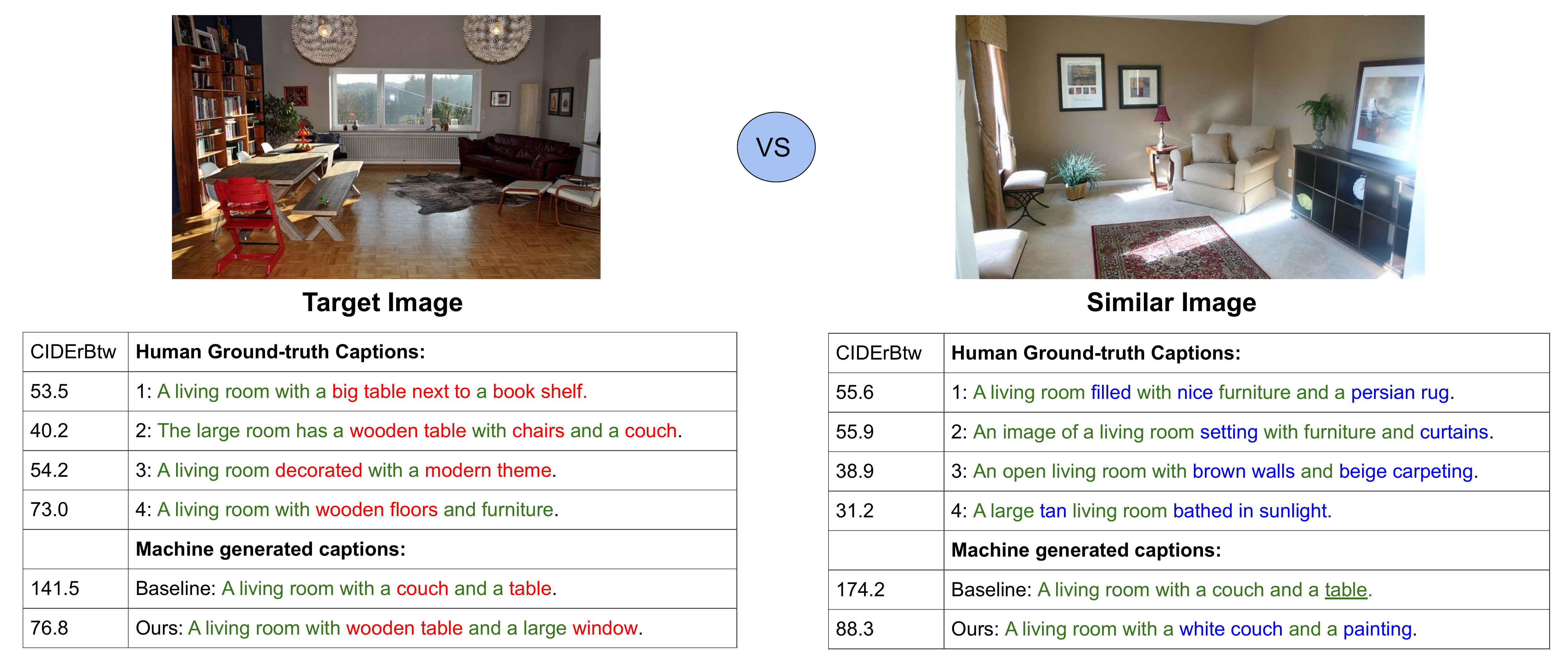}
	\caption{
		The human ground-truth captions of a target image and a semantically similar image contain both common words (highlighted in green) and distinctive words (highlighted in red for the target, and blue for the similar image). We underline the words that are irrelevant to the image. The baseline model, Transformer~\cite{vaswani2017attention} trained with MLE and SCST, generates the same caption for both images, while our model generates distinctive captions with words unique to each image. The distinctiveness is measured using CIDErBtw, the average CIDEr metric between the target caption and the GT captions of the similar images set, where lower values mean more distinctive.
	}
	\label{fig:teaser_figure}
\end{figure*}



\IEEEPARstart{I}{mage} captioning~\cite{NIC,mrnn,templatemodel,spatt,convimagecap,cnnpluscnn} attracts increasing attention from researchers in the fields of computer vision and natural language processing. It is promising in various applications such as human-computer interaction~\cite{attend2u:2018:TPAMI,vered2019joint}, information retrieval~\cite{gu2018look}, and medical image understanding~\cite{garcia2018overview}. Currently, the limitation of image captioning models is that the generated captions tend to consist of common words so that many images have similar or even the same captions (see Figure~\ref{fig:teaser_figure}). The distinctive concepts in images are ignored, which limits the wider application of image captioning since distinctive captions are needed in practice.
In terms of the quality of generated captions, \cite{liu2019generating}~summarizes four properties that encourage auto-generated captions to resemble human language: fluency, relevance, diversity, and descriptiveness. Various models and metrics have been proposed to improve the fluency and relevance of the captions so as to obtain accurate results. However, these models are poor at mimicking the inherent characteristics of human language: \textit{distinctiveness}, which refers to the specific and detailed aspects of the image that distinguish it from other similar images.

Some recent works have focused on generating more diverse captions~\cite{chen2019improving,vered2019joint,liu2019generating,sammani2020show} and discriminative captions~\cite{liu2018show,luo2018discriminability,vered2019joint}.
However, only encouraging diversity, such as using synonyms or changing word order, may not help with generating distinctive captions among multiple similar images. For instance, the human caption in Figure~\ref{fig:teaser_figure} \quotecap{an image of a living room setting with furniture and curtains} is telling the same story as \quotecap{a living room with furniture and curtains}. The two sentences have the same distinctiveness, although the first sentence is more diverse in syntax and word usage.
Discriminative captions are not always distinctive. As an example, \quotecap{a living room with a couch and a table} is discriminative since it describes the detail in the target image, but is not distinctive since it cannot describe the difference with the similar image.
In this paper, we mainly focus on promoting the \textit{distinctiveness} of image captioning, where the caption should describe the important and specific aspects of an image that can distinguish it from other similar images. To improve the distinctiveness of image captioning, we first need metrics to measure the distinctiveness of a certain caption.
The retrieval metric is widely employed in recent works to measure the distinctiveness in the evaluation and improve the distinctiveness during model training~\cite{luo2018discriminability,liu2019generating,contrastive}.
The premise of the retrieval metric is that distinctive captions should correctly retrieve the target image from a set of irrelevant images.
However, using self-retrieval in captioning models could lead to word repetition problem \cite{my-div-paper,vered2019joint}, \ie, the generated captions could repeat distinctive words, which hurts the language fluency.
Furthermore, the results may vary when choosing different retrieval models or candidate images pool. 

In this work, we propose a general metric for distinctiveness, Between-Set CIDEr (CIDErBtw), by measuring the semantic distance between one image's caption and the captions from a set of other similar images. If the caption is distinctive, \ie, captures unique concepts in its image, then it should have less overlap with its similar images set, and lead to lower CIDErBtw. To improve the distinctiveness of caption models, we also propose a training method, \textbf{D}istinctiveness via \textbf{C}omparing and \textbf{R}eweighting (DCR), which consists of three strategies that reweight the training captions and words.
In particular, we note that the human-annotated captions of each image are not equivalent on distinctiveness. Consider the example image and caption pairs shown in Figure~\ref{fig:teaser_figure}, some ground-truth captions contain more distinctive concepts (\eg, \textit{bathed in sunlight}) and detailed description that can distinguish the image from its similar image (\eg, \textit{wooden floor} and \textit{brown walls}). However, traditional training objectives such as Maximum Likelihood Estimation (MLE) and Reinforcement Learning (RL) treat each ground-truth caption and each word equally. Thus, our first strategy for improving distinctiveness is to give higher weights to the distinctive ground-truth captions and distinctive words during training. In this way, the captioning model focuses on important visual objects or properties, and generates distinctive words instead of generic ones.
 
As discussed in \cite{tan2020equalization}, the long-tail problem is often neglected in the model training. In linguistic studies, words with less frequency often contain high information entropy~\cite{shannon2001mathematical}.
However, since these low-frequency words appear less often in the training dataset, they have a low influence on model training, compared to high-frequency words.
Thus for our second strategy, we propose a weighted cross-entropy~(XE) loss to increase the training weight of low-frequency words. By encouraging the model to pay attention to low-frequency words, we thereby improve the distinctiveness of generated captions. 

Another problem of the current training process is that the common approach aims to increase the probability of the word within ground-truth captions, without deliberately suppressing the probability of other words. Although similar images often convey the same semantic meaning, there are always vital differences between them. For instance, both images in Figure~\ref{fig:teaser_figure} represent a living room, while the similar image still contains many objects that do not appear in the target image, \eg, \textit{persian rug} and \textit{white couch}. We argue that suppressing the words describing other similar images would help the model to focus on the target image itself, avoiding the generation of words that are relevant to the scenes but do not exist in the target image. To address this problem, our third strategy selects captions from the similar images set as negative samples, and encourages the model not to generate words in negative samples.

In summary, the contributions of our paper are three-fold:
\begin{itemize}
	\item We study the distinctiveness of generated captions, considering that a caption should distinguish the target image from other similar images. Similar images sets are constructed based on the similarity of the images' semantic meanings. Then a novel metric CIDErBtw is proposed to evaluate the distinctiveness of a caption within its similar image set. Experiments show that our metric aligns with human judgment for distinctiveness.
	\item We propose the Distinctiveness via Comparing and Reweighting (DCR) training method, where we give different weights to each sentence and each word. DCR method consists of three strategies: CIDErBtwStrategy guides the model to focus more on distinctive captions; The Long-Tailed Weight (LTW) strategy is integrated to highlight the rare words that contain more information; We also apply the Negative Samples (NS) strategy in contrastive learning, to further encourage the generated caption to be unique for the target image, rather than similar to other images. Experiments show that training with DCR guidance is generally applicable, and yields consistent improvement for four baseline models.
	\item  Based on the transformer network trained with XE loss, SCST (self-critical sequence training)~\cite{rennie2017self}, and DCR method, we generate distinctive captions while maintaining state-of-the-art performance according to evaluation metrics such as CIDEr and BLEU. Both automatic metrics and human evaluation demonstrate that our captions are more accurate and more distinctive. 
\end{itemize}

The preliminary version of this work was published in \cite{wang2020compare}, and has been extended in 4 aspects in this paper.  First, we further study the Long-Tailed Weight strategy that mainly works on word-level and supplements the sentence-level CIDErBtwStrategy from \cite{wang2020compare}. Second, we investigate using contrastive learning, based on word level negative samples from the similar image set, to promote the distinctiveness of generated captions. Third, we investigate the influence of different definitions of similar images sets, including different image-text retrieval methods, the size of similar images sets, and their corresponding effect on the training process and evaluating metrics.  We then collect extensive human judgments on distinctiveness, and analyze its correlation with different metrics.
Fourth, using the newly introduced methodologies, we present new experiments with the investigated DCR method on several baseline models.

The remainder of this paper is organized as follows. In Section~\ref{section:related_work}, we describe related work. We define Between-Set CIDEr in Section~\ref{sec:Between-Set-CIDEr} and propose our methodology (\ie, DCR) in Section~\ref{sec:DCR}. Next, the quantitative results, user study, and qualitative results are reported respectively in Sections~\ref{sec:experiments}, 
\ref{sec:user_study}, and \ref{sec:qualitative}. Finally, we draw a conclusion in Section~\ref{sec:conclusion}.

\section{Related work}
\label{section:related_work}
\noindent\textbf{Captioning models.} A wide range of image captioning models have been developed~\cite{NIC,mrnn,templatemodel,spatt,convimagecap,cnnpluscnn}, achieving satisfying results as measured by popular metrics, such as BLEU~\cite{bleu}, CIDEr~\cite{C} and SPICE~\cite{spice}. Generally, an image captioning model is composed of three modules: 1) visual feature extractor, 2) language generator,  and 3) the connection between vision and language. Convolutional neural networks (CNNs)~\cite{vggnet,he2016deep} are widely used as visual feature extractors. Recently, object-level features extracted by Faster-RCNN~\cite{ren2015faster} have also been introduced into captioning models~\cite{updown}, significantly improving the performance of image captioning models. \cite{yao2019hierarchy} proposed a hierarchy parsing model to fuse multi-level image features extracted by mask-RCNN~\cite{he2017mask}, which improves the performance of the baseline models. In terms of language generators, LSTM~\cite{lstm1997} and its variants are the most popular, while some works~\cite{convimagecap,cnnpluscnn} use CNNs as the decoder since LSTMs cannot be trained in parallel. More recently, transformers~\cite{vaswani2017attention,radford2019language,devlin2018bert} show improved performance in both language generation and language understanding, where the multi-head attention plays the most important role and the receptive field is much larger than CNNs. Stacking multi-head attention layers could mitigate the long-term dependency problem in LSTMs. Hence, the transformer model could handle much longer texts. For vision-language connection, attention mechanisms~\cite{spatt,rennie2017self,updown,huang2019attention} are used to reveal the co-occurrence between concepts and objects in the images. Reinforcement learning~\cite{rennie2017self,liu2018show} also guides the models to meet the metric judgment better.

\noindent\textbf{Distinctive image captioning.} Previous works~\cite{contrastive,cgan,my-div-paper} reveal that training the captioning model with MLE loss or CIDEr reward results in over-generic captions, since the captioning models try to predict an ``average'' caption that is close to all ground-truth captions. These captions lack distinctiveness, \ie, they describe images with similar semantic content using the same caption. Various recent works are related to this topic. In summary, they propose three aspects to consider: (1) {\em diversity}: describe one image with notably different expressions every time like humans~\cite{cgan}, or use rich and diverse wording~\cite{my-div-paper} to generate captions; (2) {\em discriminability}: describe an image by referring to the important and detailed aspects of the image, which is accurate, and informative~\cite{chen2018groupcap,luo2018discriminability,liu2018show,liu2019generating,vered2019joint}; (3) {\em distinctiveness}: describe the important and specific aspects of an image that can distinguish the image from other similar images~\cite{contrastive,liu2019generating}. In our paper, we focus on the last aspect, distinctiveness.

To promote diversity, two typical techniques (\ie, GAN methods and two-stage methods) have been investigated. Some works~\cite{cgan,cgan1,li2018generating,chen2019improving} employ GANs, where an evaluator distinguishes the generated captions from human annotations, encouraging the captions to be similar to human annotations. Seq-CVAE~\cite{aneja2019sequential} further improves the sentence diversity in word-wise latent space. COS-CVAE~\cite{5r_mahajan2020diverse}  splits the context and objects by factorization of the latent space, and describes the image with novel context. In the other way, the two-stage method~\cite{gu2018stack,liu2019generating,sammani2020show} first generates a simple caption and then refines it into a more diverse caption. 
StackCap~\cite{gu2018stack} proposes coarse-to-fine framework to combine multi-grained attentions; VisPara-Cap~\cite{liu2019generating} employs visual paraphrases to describe details in captions; ETN~\cite{sammani2020show} applies denoising LSTM to edit the preliminary caption.
During inference, the models with two-stage LSTM have to generate two sentences for one image, which is time-consuming. Another drawback of the two-stage method is that it cannot be trained in SCST~\cite{rennie2017self} manner, and therefore the performance based on BLEU~\cite{bleu}, CIDEr~\cite{C}, and SPICE~\cite{spice} is limited. Additionally, auxiliary information (such as where, when, and who takes the picture) could also be used to generate personalized captions~\cite{chunseong2017attend,attend2u:2018:TPAMI} to improve diversity, but many images do not have such personalized information.

Contrastive learning~\cite{contrastive,luo2018discriminability} and self-retrieval~\cite{liu2018show,vered2019joint} are introduced into captioning models to improve the discriminability of the generated captions. CL~\cite{contrastive} and DiscCap~\cite{luo2018discriminability} optimize the contrastive loss based on randomly selected unmatched image-caption pairs. However, the randomly sampled images in a batch can be easily distinguished from the target images. In contrast, instead of using random sampling, we select negative samples (NS) from the similar images set, to specifically discourage generating captions that describe the similar images. Furthermore, instead of {\em sentence-level} contrastive learning~\cite{contrastive,luo2018discriminability,liu2018show}
we focus on {\em word-level} negative samples, which more flexibly and accurately convey the distinctiveness information.
SRPL~\cite{liu2018show} employs self-retrieval reward to extend the contrastive loss, which aims at pushing the generated caption far from other images in the training batch. PSST~\cite{vered2019joint} trains the captioning model and retrieval model jointly. On the one hand, image retrieval encourages a model to generate distinctive words, while on the other hand, it hurts the accuracy and caption quality---weighing too much on image retrieval could lead a model to repeat the distinctive words~\cite{my-div-paper}.

There are some previous works~\cite{vedantam2017context,li2020context} related to generating distinctive captions. Distinctiveness requires captions to show the difference within similar images. Context-agnostic Supervision~\cite{vedantam2017context} applies suppressed loss to generate a different caption from a distractor concept.  \cite{li2020context} gives a caption for a group of images, constructing the contrastive feature to reflect the difference between images in the target group and context group.
In this paper, we encourage the generated caption to learn from its own ground-truth captions, giving higher weights to captions and words that are distinctive from other similar images, which disregards the generic captions. Thus both accuracy and distinctiveness are promoted by our method.

\noindent \textbf{Long-tail re-weighting}, \red{which uses weights to balance the loss functions of head and tail classes,} is adopted to alleviate the imbalance of long-tail distribution in image classification and object detection.  Re-weighting methods can be categorized to Class-level Weighting~\cite{2r_cui2019class,3r_wu2020distribution}  and Sample-level Weighting~\cite{1r_lin2017focal,4r_shu2019meta}. The Class-level Weighting~\cite{2r_cui2019class,3r_wu2020distribution} guides the model to pay more attention to minority categories by assigning weights to categories, based on the inverse class frequencies. On the other hand, our model focuses on each word sample and assigns the weight for the long-tail words that are more distinctive. Sample-level Weighting, \eg, focal loss~\cite{1r_lin2017focal}, determines the weight for each image by how easily the image can be correctly classified. In contrast, our training loss is combined with cross-entropy loss, and explicitly encourages the model to pay attention to the words with lower frequency in the training set.

\red{Re-sampling~\cite{SMOTE2002} is another popular direction to tackle the long-tail problem, which achieves a balanced class distribution by up-sampling the long tail classes~\cite{Zang2021FASA}. Since the supervision for image captioning is sentence-level, simply up-sampling (repeating) rare words might do harm to the sentence structure and influence the fluency of the generated captions, e.g., generating captions with repeated words. Thus we choose the re-weighting strategy to assign higher weights for the long-tail words without changing the ground truth captions structure, and leave re-sampling methods for future research.}

The long tail in the empirical distribution of the image caption dataset MSCOCO is introduced in~\cite{mrnn}. As in linguistic studies, words with less frequency often contain high information entropy~\cite{shannon2001mathematical}, and~\cite{wang2020diversity} argues that a long-tail in the word frequency distribution indicates higher diversity in the generated captions. 
\red{There are two recent works related to the long-tail phenomenon in seq-2-seq models~\cite{raunak2020long,ding2019neural}. Raunak \etal~\cite{raunak2020long} enable the model to predict the low-frequency words on the task of Neural Machine Translation. Ding \etal~\cite{ding2019neural} motivate in the opposite direction to ignore the unique information and emphasize the common words in Image Captioning models, which results in a generic caption with high accuracy as measured by CIDEr.} 
However, no prior work focuses on imposing the long-tail words in the image captioning task, since the metrics such as CIDEr and BLEU encourage the model to focus on the common words. To the best of our knowledge, we are the first to improve the distinctiveness by assigning higher weights to the long-tail words during model training.

\red{As for the techniques, Raunak \etal~\cite{raunak2020long} introduce Anti-Focal loss for penalizing low-confidence predictions, while Ding \etal~\cite{ding2019neural} calculates the word weight according to the co-occurrence between the target image and one similar image. On the contrary, our model incorporates a re-weight strategy with cross-entropy loss and assigns higher weights to the long-tail words, where the weight is based on the word frequency.}

\noindent\textbf{Metrics for distinctiveness.} Traditional metrics such as BLEU~\cite{bleu}, METEOR~\cite{M}, ROUGE-L~\cite{R}, CIDEr~\cite{C} and SPICE~\cite{spice} normally consider the overlap between a generated caption and the ground-truth captions. 
These metrics treat all ground-truth captions equally, and thus a generated caption only containing common words could obtain high scores, reflecting the statistics of human annotations.
Some works aim to generate multiple captions to cover more concepts in an image \cite{cgan,cgan1,posg,wang2019towards} and several diversity metrics are proposed, such as the number of novel captions, the number of distinctive n-grams~\cite{Dist-K}, mBLEU~\cite{cgan1}, local and global word recall~\cite{divmetric2}, and self-CIDEr~\cite{my-div-paper}. 
However, these metrics only encourage the diversity and discriminability of generated captions, and do not explicitly evaluate distinctiveness. Although generating multiple captions could cover distinctive concepts, it is difficult to summarize them into one human-like description. 

Currently, the retrieval approach is the most popular evaluation metric for distinctiveness, where a generated caption is used as the query and a pre-trained image-text embedding model, \eg, VSE++~\cite{faghri2017vse++}, is employed to rank the given images. Recall at k (R@k) is normally used to measure the distinctiveness of captions in this approach.
Ideally, a correct and distinctive caption should retrieve the image that was used to generate the caption. The drawback of retrieval-based metrics is that they are time-consuming since it uses the deep retrieval model when evaluating. Moreover, different trained models could result in different R@k.
In contrast, our proposed CIDErBtw metric for distinctiveness is fast and easy to implement, allowing it to be incorporated into various training protocols and captioning models.

\section{Between-Set CIDEr}
\label{sec:Between-Set-CIDEr}

In this paper, we aim to obtain a distinctive caption that describes the important, specific, and detailed aspects of an image. To achieve this goal, we train the captioning model to focus on important details that would distinguish the target image from semantically similar images. Our work involves two main components: the Between-Set CIDEr (CIDErBtw) value that measures the distinctiveness of an image caption from those of similar images; and several reweighting strategies for training distinctive models. We will describe the first component in this section, and the second component will be included in the next section.

The captioning model aims to generate a sentence $c^*$ to describe the semantics of the target image $I_0$. 
In the image captioning dataset, the image $I_0$ is usually provided with $N$ annotated ground-truth captions $C^0 = \{c_1^0, c_2^0, \dots, c_N^0 \}$. We first find $K$ similar images $\{I_1,I_2,\dots,I_K\}$ that are semantically similar to $I_0$, and then calculate the CIDErBtw values of $C^0$ using these similar images. CIDErBtw values indicate the distinctiveness of the captions.

\subsection{Similar images set}
\label{section:similar_image_pair}

According to the split of the training, validation, and test dataset, we measure the similarity of the target image $I_0$ to every image within the same split. Similar images sets in the training split are used when calculating the loss and the reward during training, while those in the validation and test split are used to evaluate the distinctiveness of generated captions.
For each image $I_0$ in the dataset, we find the top $K$ images $\{I_1, I_2, \dots, I_K\}$ with the highest semantic similarity to form a {\em similar images set}. 
There could be several ways to measure the similarity between two images, \eg, VSE++~\cite{faghri2017vse++}, SCAN~\cite{scan}, Vilbert~\cite{vilbert}, CLIP~\cite{radford2021learning} and CIDEr~\cite{C} similarity. We describe two effective ways in this section, \ie, VSE++~\cite{faghri2017vse++} similarity and CIDEr~\cite{C} similarity. 

\noindent\textbf{VSE++ similarity.} Given every target image, we generate its similar images set according to an image-to-caption retrieval process. We encode images and captions into a joint semantic space, and obtain similar image sets via the VSE++ retrieval model. 
Given target image $I_0$, a set of closest captions $\{{c'_1}, {c'_2}, \dots, {c'_{N_r}} \}$ in the joint space is obtained by image-to-caption retrieval, where $N_r=N (K + 1)$ to ensure that there are at least $K$+1 images in the similar images set. The top $K$ images corresponding to this captions set are considered as similar to the target image $I_0$. 
In this way, the VSE++ similarity of $I_i$ to $I_j$ is denoted as ${S_v}({I_i},{I_j})$,
\begin{align}
{S_v}({I_i},{I_j}) &= \mathop{\max }\limits_{k\in\{1,\cdots,N\}} g_r(I_i, c_k^j), \\
g_r(I_i, c_k^j) &= \frac{\phi (I_i)^T \theta (c_k^j)}{ \| \phi (I_i)  \|  \| \theta(c_k^j)  \|} \,,
\end{align}
where $\phi (\cdot)$ and $\theta (\cdot)$ are the image and caption encoders from VSE++, and $g_r(I_i, c_k^j)$ represents the retrieval score between the image $I_i$ and the $k$-th ground-truth caption of $I_j$.

\noindent\textbf{CIDEr similarity.} Since the semantics of an image is expressed in its ground-truth captions, we can directly use the similarity of ground truth captions to represent the similarity of images. In this case, the CIDEr similarity of $I_i$ to $I_j$ can be calculated by the averaged CIDEr value of their ground-truth captions as
\begin{align}
{S_c}({I_i},{I_j}) = \frac{1}{{{N^2}}}\sum\limits_{k = 1}^{N}\sum\limits_{l = 1}^{N}{g_c(c_k^i, c_l^j)} \,,
\end{align}
where $c_l^j$ is the $l$-th ground-truth caption of $I_j$, and $g_c(c_k^i, c_l^j)$ is the CIDEr value between two ground-truth captions $c_k^i$ and $c_l^j$.

\subsection{CIDErBtw}
Next, we introduce the definition of Between-set CIDEr (CIDErBtw), and why we use CIDErBtw as a metric to measure the distinctiveness of a generated caption.

\subsubsection{CIDErBtw definition}
A caption is regarded as distinctive, if it contains specific details of its target image, rather than being general enough to describe other similar images. The CIDErBtw value reflects the distinctiveness of a caption $c$ by measuring the similarity of $c$ to the ground-truth captions of similar images. Specifically, given a caption $c$ for image $I_0$, we have its similar images set $\{I_1, I_2, \dots, I_K\}$ constructed in Section~\ref{section:similar_image_pair} and their ground-truth captions $C^{(s)}=\{c^k_n\}_{n=1,k=1}^{N,K}$. We define the CIDErBtw value of the caption $c$ as
\begin{align}
CIDErBtw(c) = \frac{1}{KN}\sum\limits_{k = 1}^{K}\sum\limits_{n = 1}^{N}{g_c(c,c^k_n)}, \label{equ:CIDErBtw}
\end{align}
where $N$ is the number of ground-truth captions provided for each image, $g_c(c,c^k_n)$ represents the CIDEr~\cite{C} value between $c$ and $c^k_n$. CIDErBtw measures the averaged semantic similarity between the target caption and other ground-truth captions. A lower CIDErBtw value indicates the caption contains less commonality with the captions of its similar images.
Here we use CIDEr because it focuses more on the low-frequency words (through TF-IDF vectors) that could be more distinctive; it is efficient to compute; it is the most frequently used metric to evaluate the performance of image captioning models. Besides CIDEr, our methodology could be extended to other caption metrics to measure between-set similarity.

\subsubsection{CIDErBtw evaluation metric}
CIDEr metric measures the similarity between the generated caption $c^*$ and its ground-truth captions $C^0$, and has become an important evaluation metric in image captioning. We believe the distinctiveness should also be measured when evaluating the quality of generated captions. Thus we propose to use CIDErBtw as a complementary evaluation metric for image caption models.
We hope that the caption $c^*$ generated by the model is closer to the semantics of the target image $I_0$, while far from the semantics of other $K$ similar images $\{I_1, I_2, \dots, I_K\}$. Therefore, the $c^*$ generated by a more distinctive captioning model will have a lower CIDErBtw on the validation or test split.
CIDEtBtw requires human annotations to evaluate the generated captions, similar to other captioning metrics, \eg, CIDEr~\cite{C}, BLEU~\cite{bleu}, METEOR~\cite{M}, and ROUGE~\cite{R}. Note that for evaluation, the similar images sets are computed using the validation or test split, where appropriate. 
Another distinctiveness metric, the retrieval metric based on VSE++, also needs ground-truth captions to train the VSE++ model. Compared to the retrieval metric, our advantage is that we are fine-grained since we use similar images set to measure the distinctiveness, while VSE++ uses irrelevant images in the same batch.
Apart from acting as a metric for distinctiveness, CIDErBtw can be applied in the training strategy to improve the distinctiveness of image captioning. We will introduce the application methods in the following sections.

\section{Improving Distinctiveness via Comparing and Reweighting}
\label{sec:DCR}

In this section, we propose our training method \textbf{D}istinctiveness via \textbf{C}omparing and \textbf{R}eweighting (DCR). Our training method consists of three strategies to improve distinctiveness.  First, the {\em CIDErBtw} strategy reweights each ground truth caption according to its distinctiveness for MLE and RL training. We further improve the distinctiveness by applying a {\em Long-Tailed Weight} (LTW) strategy, which applies larger weights to long-tailed (infrequent) words. 
Finally, we propose the {\em Negative Samples} (NS) strategy, which uses negative samples from similar images set to perform contrastive learning.

\begin{figure*}[t]
	\centering
	\includegraphics[height=6.5cm]{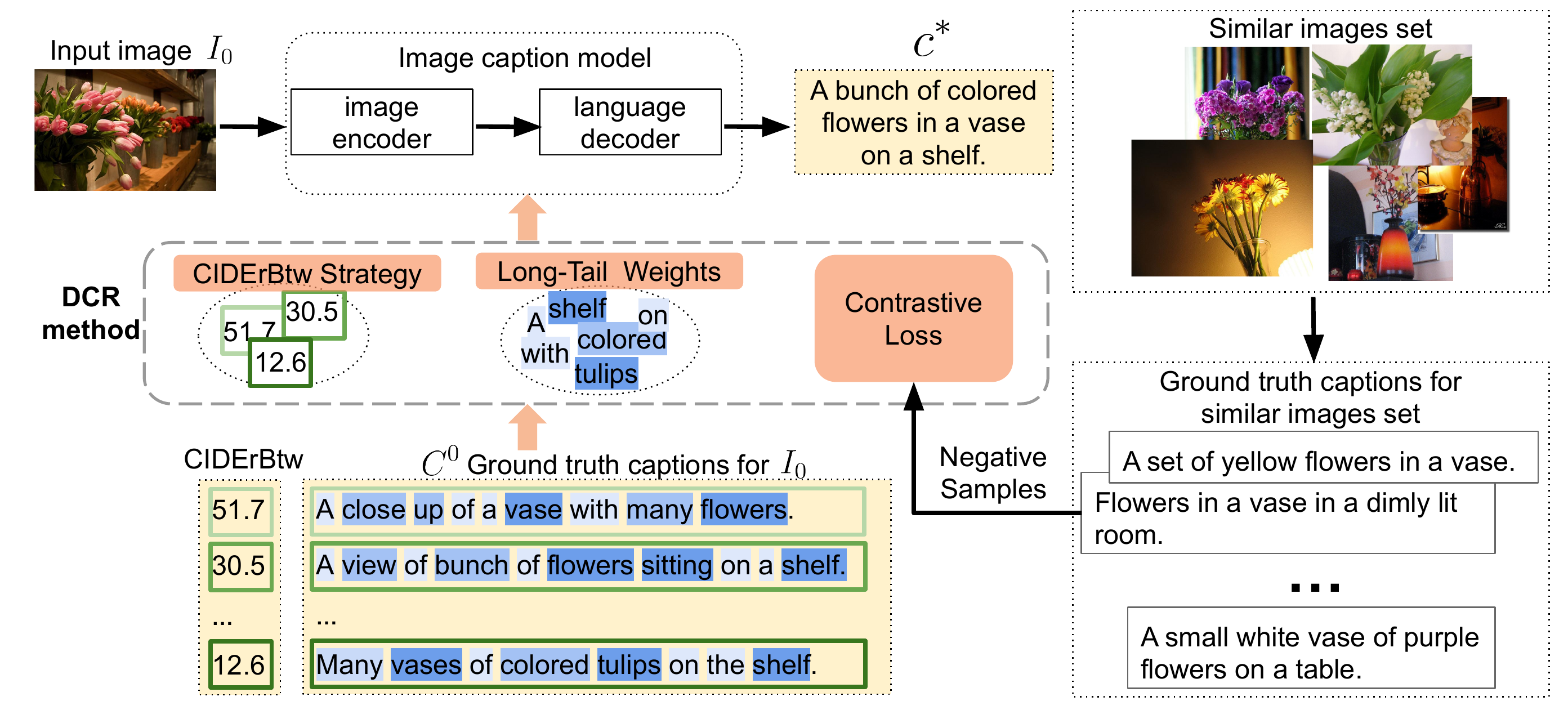}
	\caption{
	    The framework of our proposed method, Distinctiveness via Comparing and Reweighting (DCR).
	    $C^*$ and $C^0$ indicate the generated caption and the ground-truth captions.
	    The similar images set is retrieved by the input image, and CIDErBtw is computed for each ground-truth captions of the input image.
		The blue shade in the ground-truth image captions indicates word frequency, \ie, dark blue represents Long-Tail words and vice versa. The green box outlining each sentence denotes the distinctiveness, \ie, dark green represents distinctive caption, and vice versa.}
	\label{fig:structure}
\end{figure*}

\subsection{Model architecture}
Figure~\ref{fig:structure} shows the overall framework of our image captioning model, which is composed of an image encoder and a language decoder. These two modules can generate a caption $c^*$ for input image $I_0$. 

Training processes such as MLE and RL maximize the likelihood and reward for the given ground-truth captions $C^0 = \{c_1^0, c_2^0, \dots, c_N^0 \}$, which achieve state-of-the-art performance on image captioning tasks~\cite{huang2019attention, updown}. Following this idea, we apply two objective functions to update the parameters of our image captioning model, the cross-entropy (XE) loss ${\cal L}_{XE}$ and the Reinforcement Learning (RL) loss ${\cal L}_{RL}$. A hyperparameter $\alpha_l$ controls the weight of these two criteria,
\begin{align}
{\cal L} =  \alpha_l {\cal L}_{XE} + (1 - \alpha_l) {\cal L}_{RL}\,.
\end{align}
Following SCST (self-critical sequence training)~\cite{rennie2017self}, the training process of our model can be divided into two steps. The first step only trains with the cross-entropy loss ${\cal L}_{XE}$, setting $\alpha_l=1$, and the second step only trains with reinforcement learning ${\cal L}_{RL}$, setting $\alpha_l=0$. Note that our contribution lies in new training strategies that promote the distinctiveness of captions, thus our method is applicable to many baseline models trained with XE loss and RL loss.

\subsection{CIDErBtwStrategy}
\label{sec: CIDErBtw weight}

In previous models, each ground-truth caption $c_i^0$ is treated equally when calculating the XE loss and RL reward, whereas these captions might have different distinctiveness. 
In this work, we focus more on distinctive ground-truth captions by reweighting XE loss and RL reward (\ie, CIDErBtwReweight) and using CIDErBtw as a part of RL reward (\ie, CIDErBtwReward). 
For every training image $I_0$, we provide its $N$ ground-truth captions $C^0$ with different weights $W=\{w_1,w_2,\dots,w_N\}$, according to their CIDErBtw scores $V=\{v_1,v_2,\dots,v_N\}$,
\begin{align}
{v_i} &= CIDErBtw(c_i^0), \\
{w_i} &= \lambda_w  - \alpha_w \frac{{{v_i}}}{{\mathop {\max }\limits_i ({v_i})}} , \label{eq:weight}
\end{align}
where $\lambda_w$ and $\alpha_w$ are hyperparameters. Here $w_i$ indicates the contribution of the $i$-th ground-truth caption during model training. A distinctive caption $c_i^0$ will have a lower $v_i$, thus lead to higher weight $w_i$ during training.

\subsubsection{CIDErBtw weighted XE loss}
Given the words in a ground-truth caption $c_i^0 =\{d_1,d_2,\dots,d_T\}$, XE loss can be expressed as
\begin{align}
L_{XE}(c_i^0) =  - \sum\limits_{t = 1}^T \log  p_{\theta}({d_t}|d_{1:t - 1}, I_0), \label{equ:LXE_c}
\end{align}
where $p_{\theta}(d_t|d_{1:t-1}, I_0)$ denotes the probability of choosing the word $d_t$ given the word sequence ${d_1},\dots,{d_{t - 1}}$ and image $I_0$. The CIDErBtw weighted XE loss is then 
\begin{align}
{\cal L}_{XE} = \sum\limits_{i = 1}^{N} w_i L_{XE}(c_i^0)\,.
\label{equ:LXE}
\end{align}

\subsubsection{CIDErBtw weighted RL} 
Our CIDErBtwStrategy is also applied to reinforcement learning~(RL) by: 1) reweighting the reward of each ground truth captions of the target image, and 2) directly using CIDErBtw as a reward.

\noindent\textbf{CIDErBtw weighted reward.} In RL, we can reweight the reward according to the CIDErBtw to focus more on distinctive captions, resulting in a new reward, 
\begin{align}
\tilde R(c^*) = \frac{1}{N}\sum\limits_{i = 1}^{N}w_i  g_c(c^*,c^0_i),
\label{equ:newreward}
\end{align}
where $g_c(c^*,c^0_i)$ is the CIDEr value between generated $c^*$ and ground-truth $c^0_i$.

\noindent\textbf{CIDErBtwReward.} When performing RL, our CIDErBtw can also be used as a part of the reward related to distinctiveness.
We combine the CIDErBtw score  with the previous reward $\tilde R(c^*)$ and obtain the final RL reward $R(c^*)$ 
\begin{align}
\label{equ:Reinforcement}
R(c^*) &= \tilde R(c^*) - \alpha_r CIDErBtw(c^*), 
\end{align}
and RL loss ${\cal L}_{RL}$
\begin{align}
{\cal L}_{RL} &= - {\mathbb{E}_{{c^*} \sim  {p_\theta }}}[R(c^*)],
\end{align}
where $CIDErBtw(c^*)$ represents CIDErBtw score of the generated caption $c^*$ defined in (\ref{equ:CIDErBtw}), $\alpha_r$ is a hyperparameter controlling the relative contributions, and the single Monte-Carlo sampling is used as the RL policy $p_\theta$.

\subsection{Long-tailed weight strategy}
In image captioning, the frequent words are trained on sufficiently (even over-training),  while the rare words are trained on less, even though they might contain higher information entropy~\cite{shannon2001mathematical}. This problem can be addressed by giving higher weights to rare words via our Long-Tailed Weight (LTW) strategy. 
We first calculate the frequency of words in the vocabulary using the training data.
The words are then sorted in descending order by their frequency, where $f_t$ indicates the frequency index for the $t$-th word.

\begin{figure}
	\centering
	\includegraphics[width=0.9\textwidth]{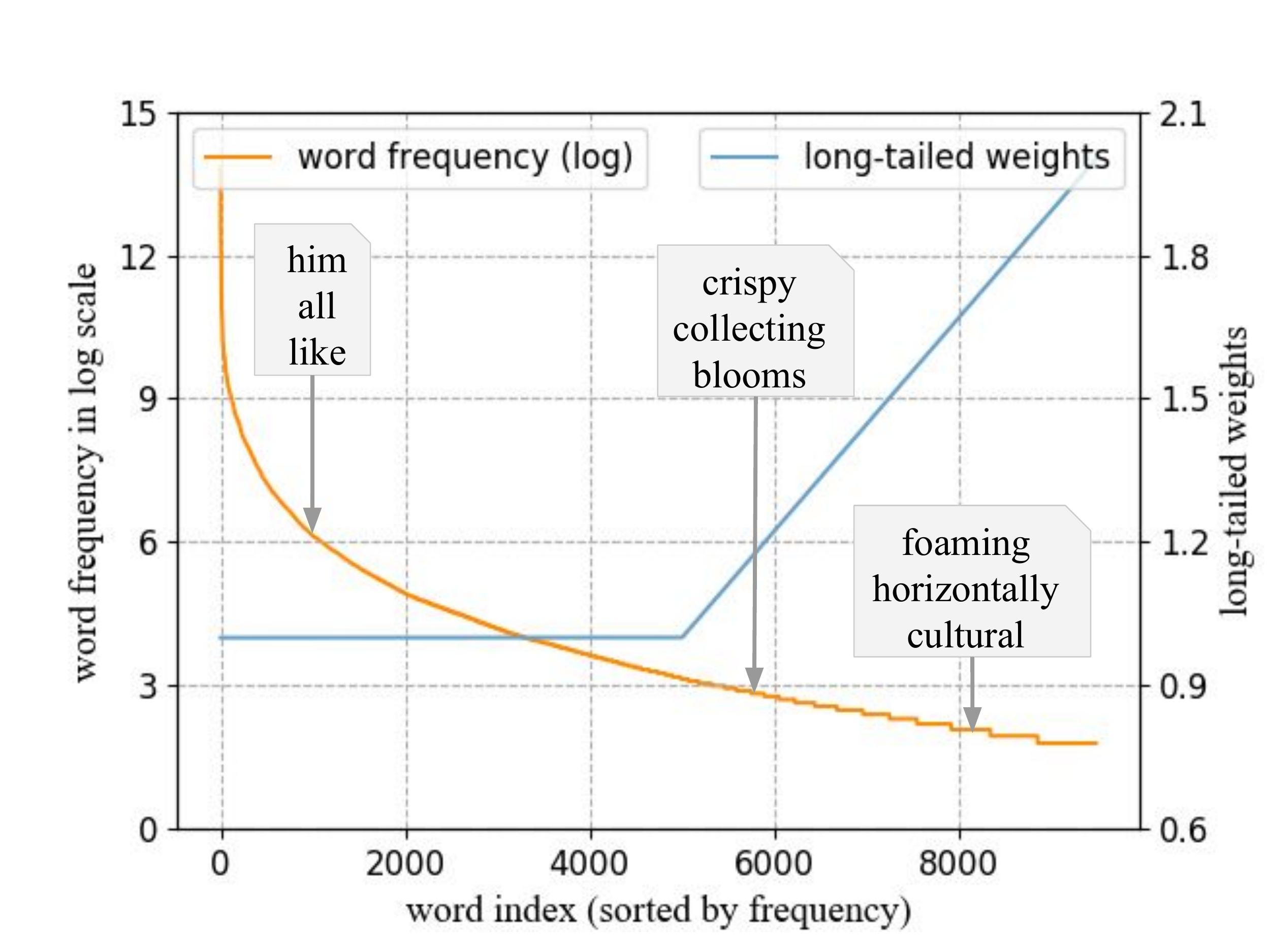}
	\caption{Word frequency distribution~(orange line) and the long-tailed weights~(blue line) assigned to each word. The beginning and ending word frequency indices to be emphasized with the long-tailed weights are $F_b$~(5,000) and $F_e$~(9,487). We show several example words with different frequencies.}
	\label{fig:word_freq_example}
\end{figure}

We propose a long-tailed weight $w_t^{\Scale[0.5]{LTW}}$ for the $t$-th word, based on its word frequency index $f_t$,
\begin{align}
w_t^{\Scale[0.5]{LTW}} = \left\{ {\begin{array}{*{20}{c}}
	{1,\qquad \quad  f_t \le {F_b}}\\
	{1 + A\frac{{f_t - {F_b}}}{{{F_e} - {F_b}}},\ {F_b} < f_t \le {F_e}}
	\end{array}} \right.
\end{align}
where $F_b$ and $F_e$ denote the beginning and ending frequency indices for words that will be emphasized, and $A$ controls the amplitude of the Long-Tailed Weights. Figure~\ref{fig:word_freq_example} displays an example plot of the word frequency distribution, the long-tailed weights, and some example words with different frequency.
The LTW strategy is applied to XE loss, so that (\ref{equ:LXE_c}) is extended to
\begin{align}
L_{XE}(c_i^0) =  - \sum\limits_{t = 1}^T w_{t}^{\Scale[0.5]{LTW}} \log  p_{\theta}({d_t}|d_{1:t - 1}, I_0).
\label{equ:LTW}
\end{align}

\subsection{Negative samples strategy}
\red{In order to suppress the generic words that also describe other similar images, 
we form a set of positive and negative caption samples for a given target image $I_0$. 
Specifically, the ground-truth captions $c_i^0$ of $I_0$ form the positive samples, while the captions $c_i^k$ from its similar image set form the {\em negative samples} (NS).
Our model is trained to give higher probabilities to positive samples, and lower probabilities to negative samples, as computed in (\ref{equ:LXE_c}).
In this case, the XE loss of (\ref{equ:LXE}) is extended to}
\begin{align}
{\cal L}_{XE} = \sum\limits_{i = 1}^{N} w_i L_{XE}(c_i^0) - \alpha_{\Scale[0.5]{NS}} \sum\limits_{k = 1}^{K}\sum\limits_{i = 1}^{N} w_i^k L_{XE}(c_i^k) \,,
\label{equ:NS}
\end{align}
\red{where $c_i^k$ is the $i$-th ground-truth caption of the $k$-th similar image, $w_i^k$ is the CIDErBtw weight of $c_i^k$, and $\alpha_{\Scale[0.5]{NS}}$ is a hyperparameter.
This loss will encourage the use of words in the positive captions, while avoid using words in the NS, thus improving distinctiveness.
}

\setlength{\tabcolsep}{2pt}
\begin{table*}[tb]
	\begin{center}
		\caption{
			Comparison of caption accuracy and distinctiveness on MSCOCO test split: (top) baseline models trained with MLE using standard or DCR weighted XE loss (denoted as DCR); (middle) models trained with SCST using standard or DCR weighted loss/reward (denoted as DCR); (bottom) SOTA methods for generating distinctive/discriminative captions.
			CIDEr, BLEU3/4, METEOR, ROUGE-L, and SPICE measure caption accuracy, while CIDErBtw and R@k measure distinctiveness.
			$\uparrow$ or $\downarrow$ show whether higher or lower scores are better for each metric. 
			CIDErBtw could not be computed for some models because the captions are not publicly available.
			Our self-retrieval results (R@k) and those of
			\cite{gu2018stack,luo2018discriminability,liu2019generating,contrastive} use the pre-trained VSE++ model and the same protocol. 
 			$\dagger$ Note that \cite{vered2019joint} reports self-retrieval results using a different retrieval model/protocol -- they use their own model for retrieval -- which makes it not directly comparable. We re-implement COS-CVAE~\cite{5r_mahajan2020diverse} on the Karpathy split~\cite{karpathy2015deep} with publicly available code.
            }
		\label{table:main_results}
		\resizebox{500pt}{130pt}{
			\begin{tabular}{l|cc|ccccc|ccc}
				\hline \hline
				Method  & CIDEr$\uparrow$  & CIDErBtw$\downarrow$ & BLEU3$\uparrow$  & BLEU4$\uparrow$  & METEOR$\uparrow$ & ROUGE-L$\uparrow$ & SPICE$\uparrow$  & R@1$\uparrow$   & R@5 $\uparrow$  & R@10$\uparrow$  \\
				\hline\hline
				FC~\cite{rennie2017self} & 97.90   & 83.35    & 41.81 & 31.58 & 25.22  & 53.34   & 17.99 & 15.44 & 40.36 & 55.08 \\
				FC+DCR (ours)  & 98.88& 83.16& 41.87& 31.77& 25.39& 53.45& 18.19& 16.18& 41.67&  56.68\\ 
				\hline
				Att2in~\cite{rennie2017self}                 & 110.04 & 83.19    & 46.36 & 35.75 & 26.79  & 56.18   & 19.91 & 17.44 & 43.88 & 58.02 \\
				Att2in+DCR (ours)  & 111.39& 82.33& 46.82&  36.27&  27.36&  56.47& 20.16&  17.88&  44.96&  59.13\\
				\hline
				UpDown~\cite{updown}                         & 111.25 & 79.46    & 45.64 & 35.93 & 27.54  & 56.24   & 20.54 & 20.10 & 47.58 & 61.92 \\
				UpDown+DCR (ours)                       & 113.60 & 78.67 & 46.64 & 36.32 & 27.78 & 56.46 & 20.67 & 21.18 & 49.84 & 63.36 \\
				\hline
				Transformer~\cite{vaswani2017attention}      & 110.13 & 80.98    & 44.80  & 34.46 & 26.98  & 55.30    & 20.18 & 21.52 & 49.88 & 64.70 \\
				Transformer+DCR (ours)                                        & 113.98 & \textbf{75.75} & 46.29 & 35.82 & 27.94 & 56.34 & 20.85 & 22.80 & 51.56 & 65.28 \\
				\hline
				\hline

				FC+SCST~\cite{rennie2017self}                & 104.43 & 90.09    & 43.10 & 31.59 & 25.46  & 54.33   & 18.67 & 11.44 & 33.16 & 48.04 \\
				FC+SCST+DCR (ours) & 104.61 & 89.31 & 43.42 & 31.42 & 25.56 & 54.56 & 18.74 & 13.16 & 36.44 & 51.14 \\
				\hline
				
				Att2in+SCST~\cite{rennie2017self}            & 117.96 & 87.40     & 47.22 & 35.31 & 27.17  & 56.92   & 20.57 & 16.00 & 41.55 & 56.66 \\
				Att2in+SCST+DCR (ours)  & 118.52& 87.11&  47.35& 35.62&  27.30&  56.98&  20.83&  17.27&  44.58&  59.14\\ 
				\hline
				
				UpDown+SCST~\cite{updown}                    & 121.94 & 86.82    & 48.82 & 36.12 & 27.95  & 57.61   & 21.29 & 18.50 & 46.34 & 61.70 \\
				UpDown+SCST+DCR (ours) & 123.10 & 86.02 & 49.83 & 37.45 & 28.36 & 58.56 & 21.75 & 21.44 & 50.44 & 65.54 \\
				\hline
				
				Transformer+SCST~\cite{vaswani2017attention} & 125.13 & 86.68    & 50.26 & 38.04 & 27.96  & 58.60    & 22.30  & 23.38 & 54.34 & 68.44 \\

				 Transformer+SCST+CIDErBtwStrategy \cite{wang2020compare} (ours)   & 127.78 & 82.74    & 50.97 & 38.52 & 29.09   & 58.82   & 22.96 & \textbf{26.46} & \textbf{57.98} & \textbf{71.28}\\	Transformer+SCST+DCR (ours) & \textbf{128.53} & 83.98 & \textbf{51.74} & \textbf{39.42} & \textbf{29.23} & \textbf{59.34} & \textbf{22.97} & 25.84 & 56.10 & 70.56 \\ \hline \hline
					COS-CVAE (consensus re-ranking)~\cite{5r_mahajan2020diverse} & 112.9  & 81.3  &47.6& 35.8   &  27.6    &  56.8  &  20.8 &    15.1    &  37.7      &    51.0  \\	Stack-Cap~\cite{gu2018stack} & 120.4          & 88.7     & 47.9 & 36.1   &   27.4    &  56.9         &   20.9       &    21.9       &    49.7      &    63.7       \\
				DiscCap~\cite{luo2018discriminability} & 120.1          & 89.2        & 48.5 & 36.1   &    27.7     &57.8           & 21.4          &   21.6       &   50.3        & 65.4          \\
				VisPara-Cap~\cite{liu2019generating} & 86.9          & -         & -  & 27.1      &  -    &-           & 21.1          &   26.3       &   57.2        & 70.8          \\
				CL-Cap~\cite{contrastive}       & 114.2           & 81.3           & 46.0           & 35.3           & 27.1           & 55.9           & 19.7              & 24.1             & 52.5             & 67.5            \\
				PSST~\cite{vered2019joint}  & 111.9           & -          & -             & 32.2           & 26.4           & 54.4           & 20.6           & 45.3$\dagger$              & 79.4$\dagger$              & 89.9$\dagger$              \\ 
				\hline \hline
			\end{tabular}
		}
	\end{center}
\end{table*}


\section{Experiments}
\label{sec:experiments}

In this section, we conduct extensive experiments to evaluate the effectiveness of our DCR method in generating distinctive captions. Note that our motivation is to generate distinctive captions as well as achieve high caption quality.

\subsection{Implementation details}

\subsubsection{Dataset and models}
We use the MSCOCO dataset~\cite{mscocodataset} with Karpathy split~\cite{karpathy2015deep}. The numbers of images are 113,287 for training, 5,000 for validation, and 5,000 for test. There are five annotated captions for each image.

For the image encoder, following Luo~\etal~\cite{luo2018discriminability}, we use two types of features in the experiments, ResNet features and spatial features. The ResNet features are extracted from ResNet-101~\cite{he2016deep}, and each image is encoded as a vector with dimension 2,048. The spatial features are extracted from the output of a Faster-RCNN~\cite{ren2015faster} following UpDown~\cite{updown}. 

Our experiments are performed on four baseline models: FC~\cite{rennie2017self}, Att2in~\cite{rennie2017self}, UpDown~\cite{updown}, and Transformer~\cite{vaswani2017attention}. FC model only uses the ResNet features, Att2in and Transformer only use the spatial features, and UpDown uses both types of features. Each model is trained using two settings: 1) Baseline model trained with ordinary MLE/SCST~\cite{rennie2017self} with standard XE/RL loss, denoted as ``\textit{model}'' and ``\textit{model}+SCST''; 
2) Baseline model trained with our DCR method, denoted as ``\textit{model}+DCR'' or ``\textit{model}+SCST+DCR''.

\subsubsection{Training details and evaluating metrics}
\red{All the hyperparameters are selected on the evaluation set of Karpathy split.} For different models, we set $\lambda_w=1.5$, $\alpha_w$ between $0.5$ to $1.0$ when reweighting the loss and the reward.
\red{We set $\alpha_r$ between $0.1$ to $0.8$ when using CIDErBtwReward}, and $0$ otherwise. We set $A=1$, $F_b=5,000$, $F_e=9,487$ (the size of vocabulary list) in LTW strategy, \red{and set $\alpha_{\Scale[0.5]{NS}}$ between 0.02 to 0.10 for the negative samples strategy.}
We use Adam~\cite{kingma2014adam} to optimize the training parameters with an initial learning rate $5\times 10^{-4}$ and a decay factor $0.8$ every three epochs. During testing, we apply  beam search with size five to generate captions. Our default similar images sets in experiments are generated using a pre-trained VSE++~\cite{faghri2017vse++} to perform the image-to-caption retrieval (see Section \ref{section:similar_image_pair}). On an NVIDIA V100, our transformer-based DCR model takes 31.5 min/epoch and 110.6 min/epoch for its two training stages, compared with 18.9 min/epoch and 102.6 min/epoch for the baseline model~\cite{vaswani2017attention}. The test-time inference has the same computation as the baseline model, which needs 17 minutes for 5,000 test images.

For evaluation, we consider two groups of metrics. The first group includes CIDEr, BLEU3/4, METEOR, ROUGE-L, and SPICE for evaluating the accuracy and quality of generated captions. The second group assesses the distinctiveness of captions, and includes our CIDErBtw metric and retrieval metrics (\ie, R@\{1,5,10\}). When calculating CIDErBtw, we use default similar images set with $K = 5$  unless otherwise specified. Thus, the CIDErBtw score measures the similarity between the generated caption and $25$ captions from the similar images set, where lower values indicating more distinctiveness. For the retrieval metrics, we follow the protocol in~\cite{liu2019generating,luo2018discriminability,contrastive}. Given a generated caption, images are retrieved in the joint semantic space of the pre-trained VSE++, with the goal to retrieve the original image. 
The retrieval metric, recall at $k$ (R@k), measures the caption-to-image retrieval performance, where a higher recall represents a better distinctiveness.

\subsection{Main results}

In this section, we present the quantitative experimental results to show the effectiveness of our DCR method at improving caption distinctiveness, and compare with state-of-the-art (SOTA).

\subsubsection{Effect of DCR}
We first present the results on the effectiveness of our DCR method on XE loss and RL reward.

\noindent\textbf{DCR on XE loss.} The main results of the model trained with XE loss and DCR are shown in Table~\ref{table:main_results}~(top). 
All baseline models obtain better performances when using DCR in the training process, which suggests that our method is widely applicable to many existing models. Specifically, our method both reduces the CIDErBtw score and improves other accuracy metrics, such as CIDEr, BLEU, etc. This shows that the generated captions become more similar to ground-truth captions, while more distinctive from other images' captions since redundancy is suppressed. We will further analyze the effect of the three strategies in DCR (\ie, CIDErBtw, LTW, and NS) separately in Section~\ref{section:Ablation}. Among the four basic models, DCR has the largest effect on Transformer~\cite{vaswani2017attention}. The reason could be the multi-head attention and larger receptive field of the Transformer allow it to extract distinctive details and context from the image. 

\noindent\textbf{DCR on RL reward.} The main results of the model trained additionally with RL (denoted as SCST) and DCR are shown in Table~\ref{table:main_results}~(middle). Our DCR method reweights the RL reward for each ground truth captions as~(\ref{equ:newreward}), and also directly uses CIDErBtw as a part of the final reward as~(\ref{equ:Reinforcement}).
Similar to the former analysis on models trained with XE loss, adding the CIDErBtw~(weighted) reward in RL improves both the distinctiveness (\ie, CIDErBtw and retrieval metrics), and the accuracy (\ie, CIDEr and BLEU) significantly. Based on Transformer+SCST, we increase the scores of the quality metrics, leading to an improvement of $3.4$ in CIDEr, $1.5$ in BLEU3, and $1.3$ in METEOR, etc. As for the distinctiveness, the CIDErBtw score of the models under these three strategies decreases from $86.68$ to $82.74$ or $83.98$. 

We observe one disadvantage of SCST that directly optimizing CIDEr reward improves the fluency of captions but also leads to common and generic words. Consistent with~\cite{my-div-paper,liu2019generating}, the baseline models trained with SCST obtain higher CIDEr but also perform worse in CIDErBtw and R@k, compared with models trained only with MLE. Optimizing the model with CIDErBtw weighted reward will relieve this problem, and the distinctiveness of captions will be promoted while maintaining or even improving the overall quality of the captions.

\subsubsection{Comparison with SOTA methods}
We list the performance of the state-of-the-art captioning models that focus on distinctiveness at the bottom of Table~\ref{table:main_results}. 
Compared to these models, our model based on Transformer+SCST generally achieves superior results in both accuracy and distinctiveness~(\ie, higher CIDEr score and retrieval score) at the same time. Specifically, Stack-Cap~\cite{gu2018stack} and DiscCap~\cite{luo2018discriminability} have lower accuracy~(CIDEr $120.4$) and less distinctiveness~(CIDErBtw 89, R@1 22), compared to our model.
By using visual paraphrases to describe details in captions, VisPara-Cap~\cite{liu2019generating} achieves high distinctiveness, which is comparable to our model, while the accuracy (CIDEr 86.9) is much worse than ours.
CL-Cap~\cite{contrastive} directly optimize the retrieval loss to improve the distinctiveness, while the R@k and CIDEr value is lower than ours.
\footnote{We could not compare distinctiveness with PSST since their captions are not publicly available, and they use a different evaluation protocol for R@k.}

\subsubsection{Reasons for improving CIDEr}
Results in Table~\ref{table:main_results} show that models trained with DCR obtain better performance for {\em both} distinctiveness metrics and accuracy metrics. Given that our training method puts higher weights on distinctive ground-truth captions, it is expected that we will obtain lower CIDErBtw and higher R@k scores. 
However, the reason why our method also improves caption accuracy (CIDEr) is less obvious, especially for SCST, which {\em directly} optimizes CIDEr using RL.
Note that CIDEr is based on the cosine similarity between TF-IDF vectors, and thus low-frequency words (with higher IDF weights) will have a higher impact on the CIDEr score. Since rare words are also distinctive, their usage in a caption should increase CIDEr.
If using distinctive words can increase CIDEr, then why does RL with CIDEr reward not use distinctive words? 
We speculate that RL gets stuck in a local minimum of models that only use frequent words because of two reasons: 1) equal weighting of an image's ground-truth captions encourages the model to predict the common words that match all captions, and 2) regularization encourages models to use smaller vocabularies -- using fewer words means less non-zero weights in the network, and lower model complexity. By reweighting the reward with CIDErBtw, more reward is obtained when using diverse words, which effectively moves the learning process out of this local minimum.

\subsection{Ablation study}

In the ablation study, we investigate the role of the three strategies of DCR, the effect of the CIDErBtw weight parameters, and consider different settings of the similar image set.

\setlength{\tabcolsep}{3pt}
\begin{table*}[t]
	\begin{center}
		\caption{Ablation study of Transformer baseline trained with different XE loss functions and reinforcement learning. ``Transformer'' means the baseline model trained with XE loss, ``+CIDErBtwReweight'' represents the CIDErBtw weighted XE loss. ``LTW'' represents Long-Tailed Weights, and ``NS'' means negative samples. These three strategies are added gradually to the baseline models. Accordingly, ``Transformer+SCST'' means the baseline model trained with XE loss and RL, and ``+CIDErBtwReweight'' represents the CIDErBtw weighted XE loss and weighted RL reward, ``+CIDErBtwReward'' represents directly optimizing CIDErBtw as a part of the RL reward.}	
		\label{table:XE_loss_guidance}
		\resizebox{500pt}{60pt}{
			\begin{tabular}{l|cccccccccc}
\hline
Method & CIDEr$\uparrow$ & CIDErBtw$\downarrow$ & BLEU3$\uparrow$ & BLEU4$\uparrow$ & METEOR$\uparrow$ & ROUGE-L$\uparrow$ & SPICE$\uparrow$ & {R@1} & {R@5} & {R@10} \\ \hline
Transformer~\cite{vaswani2017attention}      & 110.13 & 80.98    & 44.80  & 34.46 & 26.98  & 55.30    & 20.18 & 21.52 & 49.88 & 64.70 \\
\hspace{2mm}+CIDErBtwReweight \cite{wang2020compare} & 112.44 & \textbf{75.35} & 45.44 & 35.01 & 27.59 & 55.66 & 20.74 & 21.52 & 49.88 & 64.70 \\
\hspace{4mm}+LTW & 113.21 & 75.44 & 45.89 & 35.61 & 27.90 & 56.22 & \textbf{20.95} & 22.26 & 51.34 & \textbf{65.76} \\
\hspace{6mm}+NS & \textbf{113.98} & 75.75 & \textbf{46.29} & \textbf{35.82} & \textbf{27.94} & \textbf{56.34} & 20.85 & \textbf{22.80} & \textbf{51.56} & 65.28 \\ \hline

Transformer+SCST~\cite{vaswani2017attention} & 125.13 & 86.68    & 50.26 & 38.04 & 27.96  & 58.60    & 22.30  & 23.38 & 54.34 & 68.44 \\

				\hspace{1mm}+CIDErBtwReweight \cite{wang2020compare} & 128.11 & 84.70 & 51.29 & 39.00 & 29.12 & 59.24 & 22.92 & 24.46 & 55.22 & 69.02 \\
\hspace{3mm}+CIDErBtwReward  \cite{wang2020compare} & 127.78 & \textbf{82.74}    & 50.97 & 38.52 & 29.09   & 58.82   & 22.96 & 26.46 & 57.98 & 71.28 \\
\hspace{4mm}+LTW & 128.12 & 83.02 & \textbf{51.43} & \textbf{39.34} & 29.36 & \textbf{59.08} & 23.03 & 26.53 & 58.03 & 71.36 \\
\hspace{6mm}+NS & \textbf{128.24} & 82.87 & 51.31 & 39.13 & \textbf{29.42} & 58.83 & \textbf{22.75} & \textbf{26.65} & \textbf{58.12} & \textbf{71.44} \\
\hline
\end{tabular}
		}
	\end{center}
\end{table*}

\subsubsection{Three strategies of DCR}
\label{section:Ablation}

In Table~\ref{table:XE_loss_guidance}, we gradually add three strategies of DCR method (\ie, CIDErBtwStrategy, LTW, and NS) on two baseline models (\ie, Transformer and Transformer+SCST).
The CIDErBtwStrategy, which uses both CIDErBtwReweight (reweighted XE loss and RL reward) and CIDErBtwReward (additional reward of CIDErBtw value), has the most obvious effect on both accuracy and distinctiveness among the three strategies.
As shown in Table~\ref{table:XE_loss_guidance}, CIDErBtwStrategy improve the CIDEr performance (increase $2.3$ in Transformer and $2.6$ in Transformer+SCST) and CIDErBtw performance (decrease $5.6$ in Transformer and $3.9$ in Transformer+SCST) at the same time. CIDErBtwReward leads to the best retrieval metric score based on Transformer+SCST.

The Long-Tailed Weights~(LTW) and Negative samples~(NS) strategies tend to increase the CIDEr score and CIDErBtw score at the same time. LTW encourages models to generate long-tailed words, improving CIDEr performance. On the other hand, the CIDErBtw score is also increased, which suggests that these long-tailed words also appear in ground-truth captions from similar images.
The NS strategy of contrastive learning on the similar image set yields additional improvements. Compared with the CIDErBtwStrategy, one disadvantage of NS is that the CIDEr increases, while CIDErBtw fluctuates.
The reason is that there are usually some overlap words between the captions from the target image and similar images, which could confuse the model.

\begin{figure}
	\centering
	\includegraphics[height=5.6cm]{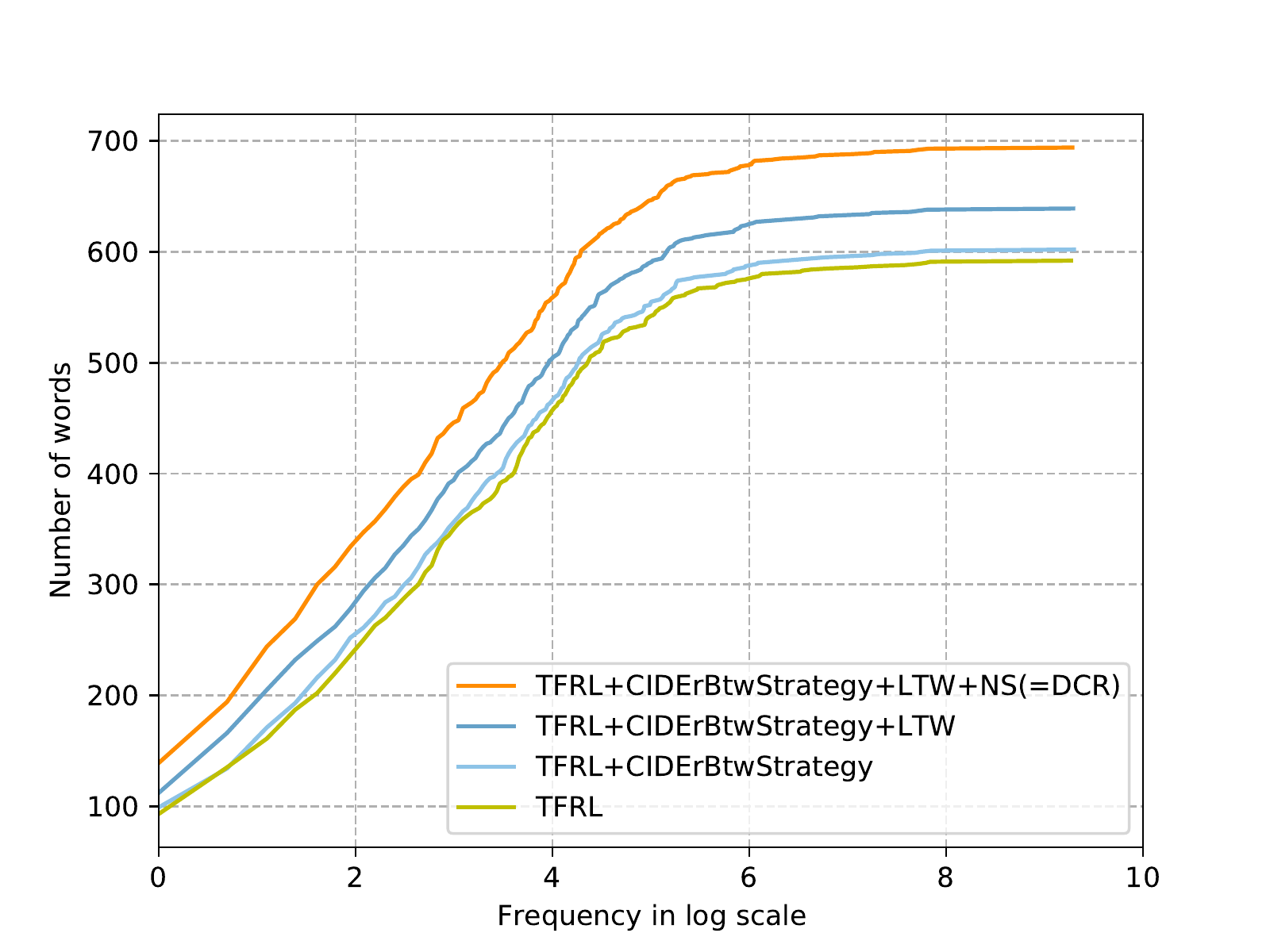}
	\caption{
		Statistics of word frequency in generated captions on the test split. The x-axis represents the word frequency $f$ in log scale, and the y-axis is the number of words with frequency smaller than $f$.   TFRL denotes Transformer+SCST. The total vocabulary sizes for the four models are 695, 640, 603, 593, which is indicated by the maximum height of each curve.
		}
	\label{fig:LTW}
\end{figure}

To show the effect of each strategy in DCR on the word usage, we compute the word frequency statistics in the generated captions on the test set. As shown in Figure~\ref{fig:LTW}, giving higher weight to the long-tailed words significantly improves the frequency of rare words in the generated captions. Using negative samples further improves both the frequency of rare words and vocabulary size. Compared to the baseline model TFRL (Transformer+SCST) with a 593-word vocabulary, our final model TFRL+DCR increases the vocabulary size by 17.2\%.

\subsubsection{Analysis of $\alpha_w$ }
 
\setlength{\tabcolsep}{2pt}
 \begin{table*}[t]
 	\begin{center}
 		\caption{The performance of our model under different CIDErBtwReweight parameter $\alpha_w$. Our model here is Transformer+SCST+CIDErBtwStrategy.} 

 		\label{table:CIDEr_recall10}
 		\resizebox{400pt}{44pt}{
 			\begin{tabular}{l|cccccccccc}
 				\hline
 				Method Condition & CIDEr$\uparrow$ & CIDErBtw$\downarrow$& BLEU3$\uparrow$ & BLEU4$\uparrow$ & METEOR$\uparrow$ & ROUGE-L$\uparrow$ & SPICE$\uparrow$ & R@1$\uparrow$  & R@5$\uparrow$  & R@10$\uparrow$ \\
 				\hline 
 				
 				Ours ($\alpha_w=1.25$)  & 125.02          & 83.29          & 49.50           & 37.11          & 28.72          & 58.37          & 22.56          & 24.54          & 54.76          & 69.34          \\
 				Ours ($\alpha_w=1.00$)  & 127.54          & 83.35          & 50.70           & 38.36          & 29.09          & 59.05          &23.04           & 25.74          & 55.90           & 70.12          \\
 				Ours ($\alpha_w=0.875$) & 127.38          & \textbf{82.48} & 50.77          & 38.34          & 29.04          & 58.77          & 22.85          & 26.34          & 57.16          & 71.14          \\
 				Ours ($\alpha_w=0.75$)  & 127.78          & 82.74          & 50.97          & 38.52          & 29.09          & 58.82          & 22.96          & \textbf{26.46} & \textbf{57.98} & \textbf{71.28} \\
 				Ours ($\alpha_w=0.50$)  & 127.61          & 83.54          & 51.22          & 38.82          & 29.10           & 59.08          &\textbf{23.09}  & 25.94          & 57.16          & 71.04          \\
 				Ours ($\alpha_w=0.25$)  & \textbf{127.96} & 83.85          & \textbf{51.33} & \textbf{38.94} & \textbf{29.12} & \textbf{59.13} & 22.90           & 25.72          & 56.04          & 70.56          \\
 				Ours ($\alpha_w=0$)    & 125.38          & 85.73          & 50.39          & 38.28          & 28.42          & 58.93          & 22.61          & 25.30           & 56.74          & 70.54       \\
 				\hline
 			\end{tabular}
 		}
 	\end{center}
 \end{table*}
 We use Transformer+SCST trained with CIDErBtwStrategy as a baseline model to evaluate the influence of $\alpha_w$. When $\alpha_w = 0$, the XE loss treats each ground truth equally. A larger $\alpha_w$ means more effect of CIDErBtw weight to the training process. As shown in Table~\ref{table:CIDEr_recall10}, the distinctiveness related metrics are first improved by increasing $\alpha_w$; however, a large $\alpha_w$ (greater than 0.875) is not conducive for learning the general language, which degrades the model performance. The model performs best in terms of distinctiveness when $\alpha_w=0.75$, while it performs best in terms of accuracy when $\alpha_w=0.25$. 
 As $\alpha_w$ increases, the training weight for those distinctive ground truth captions increases, which leads to distinctive captions, at the expense of the CIDEr score. This reflects the model's trade-off between accuracy and distinctiveness among different $\alpha_w$. Generally speaking, our model performs well when $\alpha_w$ is between 0.25 and 0.875, reflecting that our strategies are robust.  

 \red{\subsubsection{Effect of LTW} 
 The effect of LTW on accuracy and distinctiveness is shown in Table~\ref{table:XE_loss_guidance}. After introducing LTW, the CIDEr score increased (from 112.44 to 113.21 for Transformer, from 127.78 to 128.12 for Transformer+SCST), and the R@\{1,5,10\} are also increased (especially for Transformer). Since LTW does not directly optimize the CIDErBtw score, thus its effect on CIDErBtw metric is not that obvious. However, we found that LTW significantly improves the vocabulary size of the generated captions. Transformer+SCST+CIDErBtwStrategy uses 4,102 words on the MSCOCO training set, and the number increases to 4,323 (a 5.4\% increase) if LTW is additionally used.
 We perform an ablation study on the long-tail threshold $F_b$ using values $\{4000, 5000, 6000\}$. Table~\ref{table:ablation_fb} shows that models with different $F_b$ values have similar performance on accuracy (\eg~CIDEr, BLEU), while the model with $F_b=5,000$ has the best performance on distinctiveness according to R@\{1,5,10\}.
 }
 
 \begin{table*}[tb]
	\begin{center}
		\caption{
			Ablation study of $F_b$ on model Transformer+DCR.}

		\label{table:ablation_fb}
		\resizebox{370pt}{30pt}{
			\begin{tabular}{l|cc|ccccc|ccc}
				\hline 
				Method  Condition& CIDEr$\uparrow$  & CIDErBtw$\downarrow$ & BLEU3$\uparrow$  & BLEU4$\uparrow$  & METEOR$\uparrow$ & ROUGE-L$\uparrow$ & SPICE$\uparrow$  & R@1$\uparrow$   & R@5 $\uparrow$  & R@10$\uparrow$  \\
				\hline
				
				$F_b=4,000$ &113.84 & 75.92 & 46.31 & 35.76 & 27.91 & 56.47 & 20.83 & 22.72 & 51.21 & 65.18 \\
				$F_b=5,000$ & \textbf{113.98} & 75.75 & 46.29 & 35.82 & \textbf{27.94} & 56.34 & 20.85 & \textbf{22.80} & \textbf{51.56} & \textbf{65.28} \\
$F_b=6,000$ & 113.85 & 76.56 & \textbf{46.43} & \textbf{35.95} & 27.93 & \textbf{56.48} & \textbf{20.89} & 22.38 & 50.37 & 64.94 \\
without LTW & 112.82 & \textbf{75.28} & 45.67 & 35.20 & 27.61 & 55.89 & 20.75 & 21.81 & 50.16 & 64.83 \\
				\hline
			\end{tabular}
		}
	\end{center}
\end{table*}
 
 \subsubsection{Analysis of the similar images set} 
 
 Here we investigate how different settings of similar images set will influence the evaluation~(value of CIDErBtw metric), and the training process~(the performance of our CIDErBtwStrategy). We evaluate various $K$~(\ie, the number of similar images in our similar images set), as well as several methods to construct the similar images set (including retrieval similarity and CIDEr similarity).
 We use CBV to denote CIDErBtw calculated from similar images set constructed by VSE++, and use CBC to denote CIDErBtw calculated from similar images set constructed by CIDEr similarity. \red{We use Transformer+SCST trained with CIDErBtwReweight~\cite{wang2020compare} as the baseline.}
 
 \noindent\textbf{CIDErBtw metric evaluation.} We first evaluate how different numbers of images will affect the CIDErBtw value. As shown in Table~\ref{table:different_K} (middle), CBV(K) decreases as $K$ increases. The reason is that the semantic similarity between target image $I_0$ and the $k$-th similar image $I_k$ will decrease as the $k$ increases, which results in a smaller CIDEr value between their ground-truth captions. Although the CIDErBtw score varies for different $K$, the score difference between the two models is stable, which indicates the stability of our CIDErBtw metric.
 
 Next we compare the CIDErBtw metric constructed by VSE++~(CBV) and CIDEr~(CBC) similarity. The distinctiveness evaluation result using CBC is also shown in Table~\ref{table:different_K} (right).
 In general, CBC scores are higher than CBV scores, because CBC finds images that share almost the same GT caption with the target image. However, our aim is not to compare the slight differences between the two captions that are almost the same. We intend to calculate CIDErBtw among images with similar semantics, but not with exactly the same captions.  Thus, CBV is a more appropriate choice for measuring distinctiveness.

 \noindent \textbf{Effect of the similar image number K.}
We next study the effect of using different numbers of images $K$ in the similar image sets.  We train our model using similar image set with different $K$, and the results are presented in Table 5. Using $K$ from 3 to 7 yields similar results (CIDEr from 127.96 to 128.12). Increasing the number of similar image sets to $K=10$ cannot improve the results further, since $7$ similar images already well depict the distinctiveness of each ground truth caption. \red{The reason is that higher K introduces similar images that are far from the target image, \eg, images that have different semantic meanings. The captions for those ``far-away'' similar images may be too different from all the ground truth captions of target image, thus they cannot well evaluate the distinctiveness of target image captions, which introduces noise to our model.} On the other hand, reducing $K$ to 1 results in CIDEr of $125.45$, which is only slightly better than the baseline (Transformer+SCST) of $125.13$. The top-1 similar image usually shares too much semantics with the target, and thus cannot well distinguish between different ground truth captions.

\noindent \textbf{Effect of the image-text retrieval model.} We next perform an ablation study on using different image-text retrieval models to construct the similar image sets for training. Specifically, we consider four additional retrieval methods: CIDEr similarity, SCAN~\cite{scan}, Vilbert~\cite{vilbert}, \red{CLIP~\cite{radford2021learning},} and randomly selecting images. The results are presented in Table 5~(middle), and indicate that different image-text retrieval models work on par with each other. Retrieving similar images with VSE++~\cite{faghri2017vse++}, \ie, CBV(7), yields the best CIDEr of $128.12$, \red{while retrieving similar images with SCAN, Vilbert, and CLIP achieves on par CIDEr value at $127.98$, $128.09$, and $127.81$ respectively}. The results indicate that our model performance is not sensitive to the image-text retrieval method as long as images in the similar image set have similar semantic meaning. 

\red{Another method to reweight each sentence is to measure the CIDErBtw of each caption with respect to the other four captions of the target image (denoted as CB-GT-Caption). In essence this is selecting the most diverse ground-truth caption for the image. This model only achieves on par accuracy and distinctiveness with the CBV(1) model.} \red{We also directly compare the images' feature similarity with CLIP model and generate similar image sets (indicated by CB-Image). CB-Image method does not achieve as good performance as those retrieval methods.} 

The reason that CBC weighting performs slightly worse~(\ie, $127.54$ with CIDEr) is that the similar images retrieved by CIDEr share almost the same GT captions as the target image.  Thus, using CBC as loss/reward weight will reduce the influence of all the ground-truth captions. In contrast, CBV finds similar images with closer semantic meaning, instead of an image with the same ground-truth captions as CBC does.  Since CBV(5) is a more reasonable approach for the similar image set, we make CBV(5) the default setting in our experiments. On the other hand, selecting random images to form similar images sets harms the model performance, resulting in worse results than the baseline model without CIDErBtw training strategy -- Transformer+SCST+CB-Random obtains  $123.67$ with CIDEr compared to $125.13$ for the baseline transformer+SCST. This indicates that weighting the ground-truth captions for the target image according to its similarity with random image captions will introduce noise to the model and harm the training process.

 \begin{table*}
 	\resizebox{340pt}{18pt}{
 		\begin{tabular}{l|c|ccccc|c}
 			\hline
 			\setlength{\tabcolsep}{0.3mm}
 			Models           & CIDEr$\uparrow$  & CBV(1)$\downarrow$ & CBV(5)$\downarrow$ & CBV(7)$\downarrow$ & CBV(10)$\downarrow$ & CBV(15)$\downarrow$ & CBC(5)$\downarrow$\\ \hline
 			Transformer+SCST          & 125.13 & 93.10         & 86.68         & 83.75         & 79.89          & 74.08          & 135.35                        \\ 
 			Transformer+SCST+CBV(5) & 128.11 & 92.23        & 84.70         & 80.82         & 77.01          & 71.08          & 135.16                        \\ \hline
 		\end{tabular}
 		\caption{The evaluation results for different similar images settings. We use Transformer+SCST as the baseline model. Here CBV($K$) represents the CIDErBtw metric with $K$ similar images constructed by VSE++ similarity. CBC($K$) represents the CIDErBtw metric with $K$ similar images retrieved by CIDEr similarity. \red{Note that these models are trained under CBV(5) when using Transformer+SCST trained with CIDErBtwReweight.}}
 		\label{table:different_K}
 	}
 \end{table*}
 
 \begin{table*}
 	\resizebox{400pt}{75pt}{
 		\begin{tabular}{cc|cccccccc}
\hline 
\multicolumn{2}{c|}{Method} &  CIDEr$\uparrow$  & CBV(5)$\downarrow$ & CBC(5)$\downarrow$ & BLEU3$\uparrow$ & BLEU4$\uparrow$ & METEOR$\uparrow$ & ROUGE-L$\uparrow$ & SPICE$\uparrow$ \\
\hline
\multirow{12}{*}{transformer+SCST+} & CBV(1) & 125.45 & 86.24 & 140.39 & 50.43 & 38.24 & 28.18 & 58.84 & 22.46 \\
 & CBV(3) & 127.96 & 84.76 & 136.83 & 50.64 & 38.84 & 28.46 & 58.73 & 22.64 \\
 & CBV(5) & 128.11 & \textbf{84.70} & 135.16 & \textbf{51.29} & 39.00 & \textbf{29.12} & \textbf{59.24} & 22.92 \\
 & CBV(7) & \textbf{128.12} & 84.89 & 135.83 & 51.01 & 38.72 & 28.89 & 59.09 & 22.67 \\
 & CBV(10) & 127.69 & 85.73 & 137.42 & 50.97 & 38.68 & 29.01 & 59.11 & 22.82 \\
   \cline{2-10}
   & CB-Image & 126.23 & 85.92 & 137.14 & 50.71 & 38.46 & 28.48 & 58.76 & 22.54 \\
 		& CB-GT-Caption & 125.50 & 86.21 & 137.84 & 50.39 & 38.19 & 28.41 & 58.63 & 22.41\\
   
   \cline{2-10}
 & CBC & 127.54 & 84.81 & \textbf{134.07} & 51.26 & 38.89 & 29.00 & 58.93 & \textbf{22.93} \\
 & CB-SCAN~\cite{scan} & 127.98 & 84.92 & 135.53 & 51.09 & 38.95 & 28.97 & 59.19 & 22.52 \\
 & CB-Vilbert~\cite{vilbert} & 128.09 & 84.83 & 135.83 & 51.17 & \textbf{39.03} & 29.05 & 59.04 & 22.72 \\
& CB-CLIP~\cite{radford2021learning} & 127.81 & 84.72 & 135.02 & 50.94 & 38.66 & 28.93 & 58.87 & 22.65 \\
 & CB-Random & 123.67 & 86.42 & 139.75 & 49.79 & 37.59 & 27.95 & 58.57 & 22.25 \\
 \hline
\multicolumn{2}{c|}{transformer+SCST}   & 125.13 &  86.68  & 135.35 & 50.26 & 38.04 & 27.96  & 58.60    & 22.30  \\
 \hline 
\end{tabular}
 		\caption{Top (Row 1 to 5): the models trained under different similar image set size. Here CBV($K$) represents the CIDErBtw under $K$ similar images retrieved by VSE++ Retrieval similarity. Middle (Row 6 to 7): CB-Image indicates similar images set constructed by images' feature similarity. CB-GT-Caption indicates using the target image itself to reweight each caption. Bottom  (Row 8 to 12): the models trained with similar image sets constructed from four different image-text retrieval methods, CIDEr similarity~(CBC), SCAN~\cite{scan}~(CB-SCAN), Vilbert~\cite{vilbert}~(CB-Vilbert), CLIP~\cite{radford2021learning}~(CB-CLIP) and random sampling~(CB-Random). The number of images in the similar images set is $K=5$. Note that ``CBV(5)'' after the model name means training strategies, while in the table header it means evaluation metric. These models are Transformer+SCST trained with CIDErBtwReweight under different similar image sets.}
 		\label{table:different_similarity}
 	}
 \end{table*}

\section{User Study}
\label{sec:user_study}
We conduct two user studies to evaluate the quality of generated sentences, and verify the consistency between the proposed distinctiveness metric and human judgment. The influence of our DCR method is evaluated in the first user study. The comparison between our model and the state-of-the-art distinctiveness models is conducted in the second user study. In total, twenty users participated in the experiments.

\subsection{Influence of DCR method}
We perform the experiment on two baseline captioning models: UpDown~\cite{updown} and  Transformer~\cite{vaswani2017attention} trained by SCST. Then we test the quality of the captions of both models trained with and without our DCR method, where we randomly sample 50 similar images pair from the test split. This user study includes two parts: image retrieval game and quality judgment.

\subsubsection{Image retrieval game}
In the image retrieval game, to evaluate the distinctiveness of generated captions, we ask the user to distinguish the target image from a similar image, according to the caption that describes the target image. The target-similar image pair are constructed by VSE++~(see Section~\ref{section:similar_image_pair}). Following the protocol in \cite{luo2018discriminability}, the game involves displaying the target image, a similar image with the same semantic meaning, and a generated caption describing the target image. The users are asked to choose the image that more closely matches the caption. 
 
\begin{table}[!h]
	\begin{center}
		\caption{
			User study on image retrieval to assess caption distinctiveness. Our models trained with DCR generated more distinctive captions, enabling the user to more accurately select the correct image, compared with the baselines (2-sample z-test on proportions,  $^*$ p$<$0.05, $^{**}$ p$<$0.01).
		}
		\begin{tabular}{c|c}
			\hline 
			Method            & image retrieval game   \\
			\hline
			UpDown+SCST            & 68.7\%  
			\\
			UpDown+SCST+DCR & \textbf{73.9\%} $^{**}$\\
			\hline
			Transformer+SCST  & 75.2\% 
			\\
			Transformer+SCST+DCR       & \textbf{79.8\%} $^{*}$        \\
			\hline 
		\end{tabular}
		\label{tab:user_study}
	\end{center}
\end{table}

\begin{figure}
	\centering
	\includegraphics[width=7 cm]{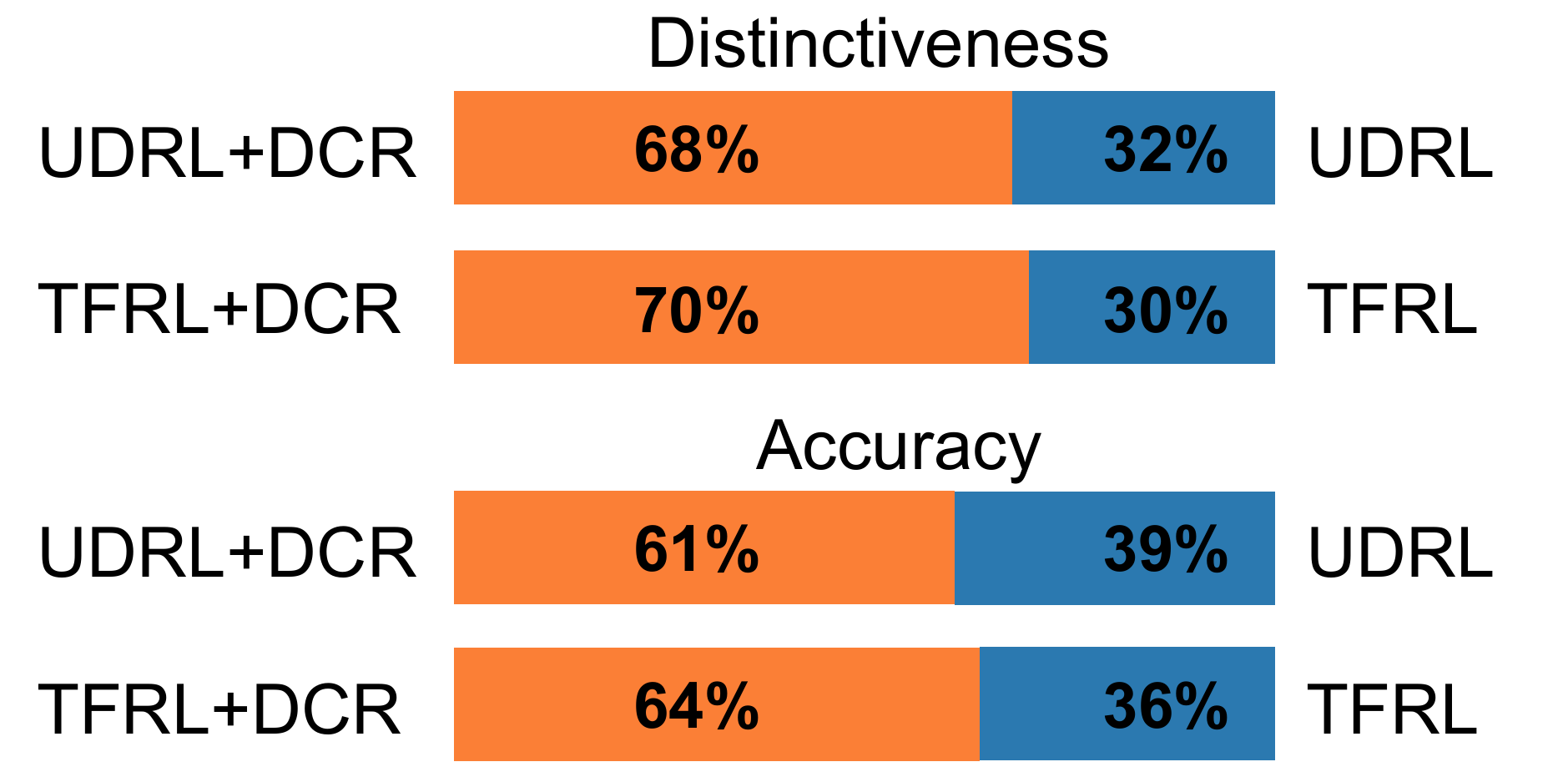}
	\caption{User study comparing captions generated from models trained with and without our DCR method. Here UDRL and TFRL denote UpDown+SCST and Transformer+SCST for short. Users selected our models trained with DCR more frequently when assessing accuracy and distinctiveness (Chi-Square test, p$<$0.001 for each pair).
	}
	\label{fig:user_study}
\end{figure}

The results for the image retrieval game are shown in Table~\ref{tab:user_study}. Compared to the baseline model, our DCR method increases the accuracy of human image retrieval performance by $5.2\%$~(on UpDown+SCST) and $4.6\%$~(on Transformer+SCST). The results of user study are consistent with the automatic image retrieval results (R@k), which indicates that our model generates captions with more distinctive descriptions, which can help both machines and humans to retrieve the target image. 

\subsubsection{Quality judgment}
In the quality judgment test, we compare captions generated from a baseline model trained with and without DCR. In each trial, an image and two captions are displayed, and the user is asked to choose the better caption with respect to two criteria: distinctiveness and accuracy. The interface can be found in the supplementary material.

Figure~\ref{fig:user_study} shows the results for the quality judgment test. 
From the human perspective, captions from our models are more distinctive than the baseline models: $68\%$~(ours) vs.~$32\%$~(UDRL baseline); $70\%$~(ours) vs.~$30\%$~(TFRL baseline). The human judgment also prefers captions from our models over the baseline on accuracy:
$61\%$~(ours) vs.~$39\%$~(UDRL baseline); $64\%$~(ours) vs.~$36\%$~(TFRL baseline). Again this is consistent with the observations from the machine-based metrics (CIDErBtw and CIDEr).

\subsection{Comparision with SOTA methods}
In this section, we compare our method and other state-of-the-art distinctive methods~\cite{contrastive,gu2018stack,luo2018discriminability} through a user study. Similar to the first user study above, we also perform image retrieval game and quality judgment test. In this experiment, we randomly sampled $50$ images from the test split. Firstly, we performed an image retrieval game with captions generated by these models and report the accuracy in Table~\ref{table:SP_user_study}. The image retrieval accuracy is consistent with other metrics reported in Table~\ref{table:main_results}, where our method achieves the most distinctive result, higher than the second-best, CL-Cap, by a large margin ($3.8\%$ in retrieval accuracy).
In the quality judgment part, we compare captions generated by four methods, Stack-Cap~\cite{gu2018stack}, DiscCap~\cite{luo2018discriminability}, CL-Cap~\cite{contrastive}, and our model~(Transformer trained with SCST and DCR method). In each trial, an image and four captions are displayed, and eight users are asked to rate these captions from the 1-5 scale (higher is better) with respect to two criteria (distinctiveness and accuracy). The results are shown in Table~\ref{table:SP_user_study}. Our method achieves higher scores in both distinctiveness and accuracy, which is consistent with the automatic evaluation reported in Table~\ref{table:main_results}.

\setlength{\tabcolsep}{2pt}
\begin{table}[tb]
	\begin{center}
		\caption{User study results on image retrieval and caption rating. Here TFRL denotes Transformer+SCST for short. Compared with SOTA methods, our models achieve higher image retrieval accuracy and rating scores (in statistical significant test with all baselines, $^{\star}$ indicates 2-sample z-test on proportions with p$<$0.05, $^{\star\star}$ indicates paired t-test with p$<$0.05).}
		\label{table:SP_user_study}
		\resizebox{250pt}{30pt}{
			\begin{tabular}{c|c|c|c}
				\hline 
				Method            & image retrieval & distinctiveness & accuracy   \\
				\hline
				Stack-Cap~\cite{gu2018stack}  &  72.6\%  & 3.34 & 3.33\\
				DiscCap~\cite{luo2018discriminability} & 73.9\% &3.37 &3.41\\
				CL-Cap~\cite{contrastive}  & 75.8\% &3.41 &3.12\\
				TFRL+DCR(ours) & \textbf{79.6\%} $^{\star}$  & \textbf{3.57}$^{\star\star}$ & \textbf{3.45}$^{\star\star}$  \\
				\hline 
			\end{tabular}
		}
	\end{center}
\end{table}

 \subsection{CIDErBtw as a metric}
 \label{section:CIDErBtw_metric}
 In the third user study, we investigate the suitability of the CIDErBtw metric for distinctiveness. It is believed that we should evaluate the generated captions in different aspects (\eg, accuracy, fluency, and distinctiveness, etc.), and none of the existing metrics can well evaluate captions in all aspects. The CIDErBtw metric works to complement the existing metrics, as it can measure the distinctiveness of captions. Note that the CIDErBtw metric evaluates the n-gram distance between the generated caption and a set of captions from other similar images.
 When evaluating the quality of a generated caption, we should consider both CIDEr and CIDErBtw, since the former metric evaluates the similarity to all ground truth captions (accuracy), while the latter represents how different are the generated captions compared to other captions from similar images (\ie, distinctiveness).
 
 \begin{table}
 	\resizebox{250pt}{34pt}{
 		\begin{tabular}{c|cc|cc|cc}
 			\hline
 			\multirow{2}{*}{Compared models} & \multicolumn{2}{c|}{Pearson} & \multicolumn{2}{c|}{Spearman} & \multicolumn{2}{c}{Kendall} \\ \cline{2-7} 
 			& VR & CB    & VR & CB      & VR & CB       \\ \hline
 			DiscCap \& StackCap                & 0.143         & 0.172          & 0.118         & 0.167          & 0.094         & 0.177          \\
 			TFRL (w \& w/o DCR)             & 0.016         & 0.106          & 0.246         & 0.208          & 0.213         & 0.240          \\ TFRL+DCR (First \& Second) & 0.202         & 0.273          & 0.263         & 0.298          & 0.217         & 0.285          \\
 			\hline
 			Overall correlation       & 0.134         & \textbf{0.187} & 0.215         & \textbf{0.223} & 0.169         & \textbf{0.176} \\ \hline
 		\end{tabular}
 		\caption{The correlation between human judgment and automatic metrics, \ie, VSE++ Recall (VR) and CIDErBtw metric (CB). We use three typical indicators (\ie, Pearson, Spearman, and Kendall) to compare the correlation between human judgment and the automatic metric on distinctiveness. The user study is conducted for three model pairs, each with 200 caption pairs. The overall correlation is computed on 600 numbers (3 conditions, 200 for each). 
 		TFRL denotes Transformer+SCST for short, ``w \& w/o DCR'' means the model trained with and without DCR method, and  ``First \& Second'' means the first and second captions in beam search.
 		}
 		\label{table:correlation}
 	}
 \end{table}

 \begin{figure*}[h]
	\centering
	\includegraphics[width=16cm]{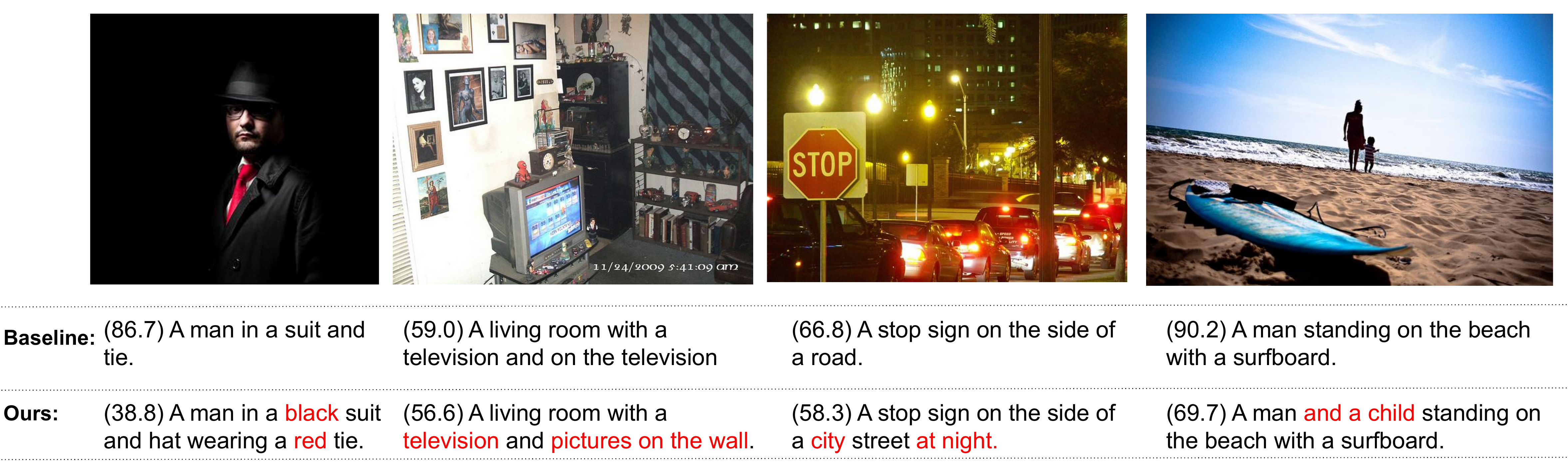}
	\caption{Example captions from the baseline model and our model. The distinctive words are highlighted. The number in parenthesis is the CIDErBtw score, with lower values meaning more distinctive.
	}
	\label{fig:qualitative_1}
\end{figure*}

\begin{figure*}[h]
	\centering
	\includegraphics[width=18cm]{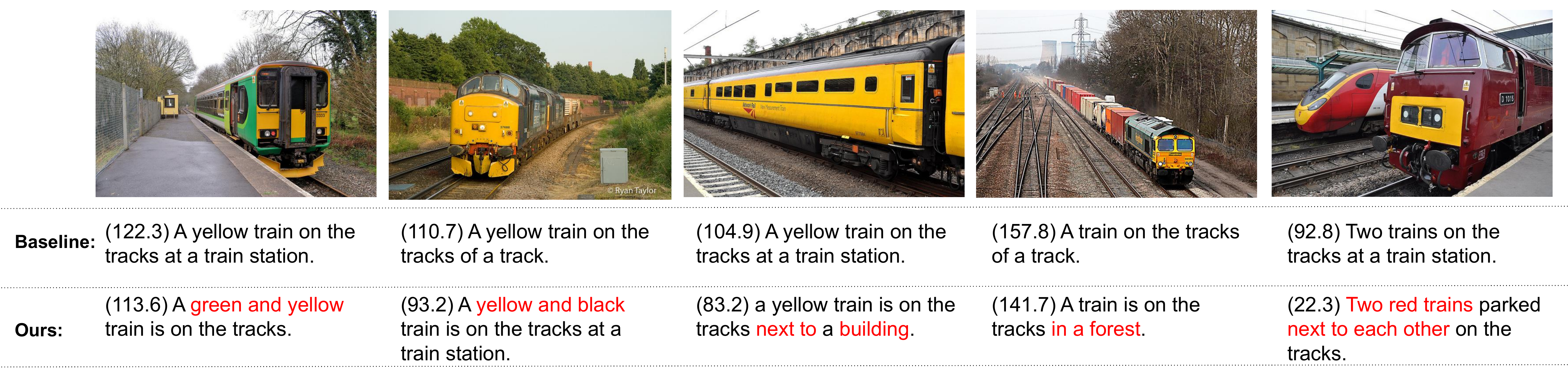}
	\caption{Example captions for a set of similar images. The distinctive words are highlighted in red.
	}
	\label{fig:qualitative_2} 
\end{figure*} 
 
 Following~\cite{wang2020diversity}, we perform a user study to compare the correlations of CIDErBtw metric (CB) and VSE++ Recall (VR) to human judgment. 
 To show the generalization of our metric to varying degrees of model comparisons,  
 we run experiments on captions from three model pairs:
 1) two captions from different models (DiscCap and StackCap);
 2) two captions from similar models trained with two different conditions (Transformer+SCST models trained with and without DCR); 
3) two captions from the same model (the 1st and 2nd results from beam search on Transformer+SCST+DCR).
The first pair reflects a coarse comparison between completely different models, while the third pair is a fine-grained comparison within the same model. 

For each of the three model pairs, we randomly sampled 200 images from the test split for each of the three model pairs.
For each caption pair generated from an image, six participants were asked to score the more distinctive caption (scoring from -6 to 6; the value -6 means the first caption is absolutely more distinctive than the second in the caption pair, and vice versa). The automatic metric is calculated by subscribing to the value generated by CB (or VR) for two captions. Finally, the correlation was calculated between human judgment scores and the automatic metric with three correlation indicators, Pearson, Spearman, and Kendall.
 
 The user study results are shown in 
 Table~\ref{table:correlation}. Our CIDErBtw metric has a higher overall correlation with human judgment compared to VR. Furthermore, our metric has a higher correlation than VR on almost all the model settings and under three indicators, which indicates that CIDErBtw is more suitable as a distinctiveness metric, with respect to human judgment.

\section{Qualitative Results}
\label{sec:qualitative}

In this section, we show qualitative results for the baseline model Transformer+SCST, and our model Transformer+SCST+DCR in Figure~\ref{fig:qualitative_1}.
The baseline model generates captions that accurately describe the main object, but are quite generic and monotonous, while captions generated by our model are more distinctive in the following aspects. Our captions describe more properties of the main object, such as \quotecap{black suit}, \quotecap{red tie} and \quotecap{a man and a child}. Our captions also describe backgrounds that are distinctive, such as \quotecap{pictures on the wall} and \quotecap{city street at night}. 
Intuitively, in order to increase a caption's distinctiveness, the model should focus on more properties that would distinguish the image from others, such as properties of the most important object, color, numbers, or other objects/background in the image. Our method generates captions that focus on more of these aspects and generates accurate results.

In order to show the distinctiveness of our model, we present a  similar images set with the same semantic meaning in Figure~\ref{fig:qualitative_2}.
The baseline model generates captions that follow templates like \quotecap{train on the track} or \quotecap{at a train station}. Although the captions are correct, it is difficult to tell the images apart according to the captions. Our model generates more distinctive captions by enriching the description and mentioning the environment around it. For example, the colors such as \quotecap{green and yellow} and \quotecap{yellow and black} distinguish the first two images. The background environment such as \quotecap{
next to a building} and the \quotecap{in a forest} is also observed, which is quite encouraging. Furthermore, our model is more sensitive to the relative positions of objects, such as \quotecap{next to each other on the tracks}. 
More details can be found in the Appendices.

\section{Conclusion}
\label{sec:conclusion}
In this paper, we consider an important property, \textit{distinctiveness} of image captioning models, and proposed a metric CIDErBtw to evaluate distinctiveness, which can be calculated quickly and easily implemented. We found that human-annotated captions for each image vary in distinctiveness based on CIDErBtw.
To improve the distinctiveness of generated captions, we developed a novel training method, Distinctiveness via Comparing and Reweighting (DCR), consisting of three strategies.
In the first strategy, each human ground-truth annotation is assigned a weight based on its distinctiveness computed by CIDErBtw. Thus, during training the model pays more attention to the captions that are more distinctive. We also consider using CIDErBtw directly as part of the reward in RL. 
In the second strategy, we propose Long-Tailed Weights (LTW) to giving higher weights to rare long-tailed words that might contain more information. 
In the third strategy, to further promote the distinctiveness of generated captions, we take the captions from similar images set as negative samples in contrastive learning. In extensive experiments, we showed that our method is widely applicable to many captioning models. Experimental results demonstrate that our training strategy is able to improve both accuracy and distinctiveness, achieving state-of-the-art performance on CIDEr, CIDErBtw, and retrieval metrics (R@k). 

Distinctive image captioning requires generated captions to be both accurate and distinct. In terms of problem definition, semantically closer images will make the comparison more valuable. In terms of model training, captioning models need to be clearly aware of the differences between these similar images. In terms of evaluation, reasonable metrics are needed to ensure achieving real distinctive models. Our CIDErBtw metric and DCR training method aim to meet these requirements. In future work, we will apply distinctive methods on other tasks (\eg, Visual Question Answering and Object Detection) to suppress generic performance.

\section*{Acknowledgments}

This work was supported by grants from the Research Grants Council of the Hong Kong Special Administrative Region, China (Project No. CityU 11212518 and CityU 11215820) and from the City University of Hong Kong (Strategic Research Grant No. 7005218).

\ifCLASSOPTIONcaptionsoff
  \newpage
\fi

\bibliographystyle{IEEEtran}
\bibliography{mainbib}

%


\begin{IEEEbiography}[{\includegraphics[width=1in,height=1.25in,clip,keepaspectratio]{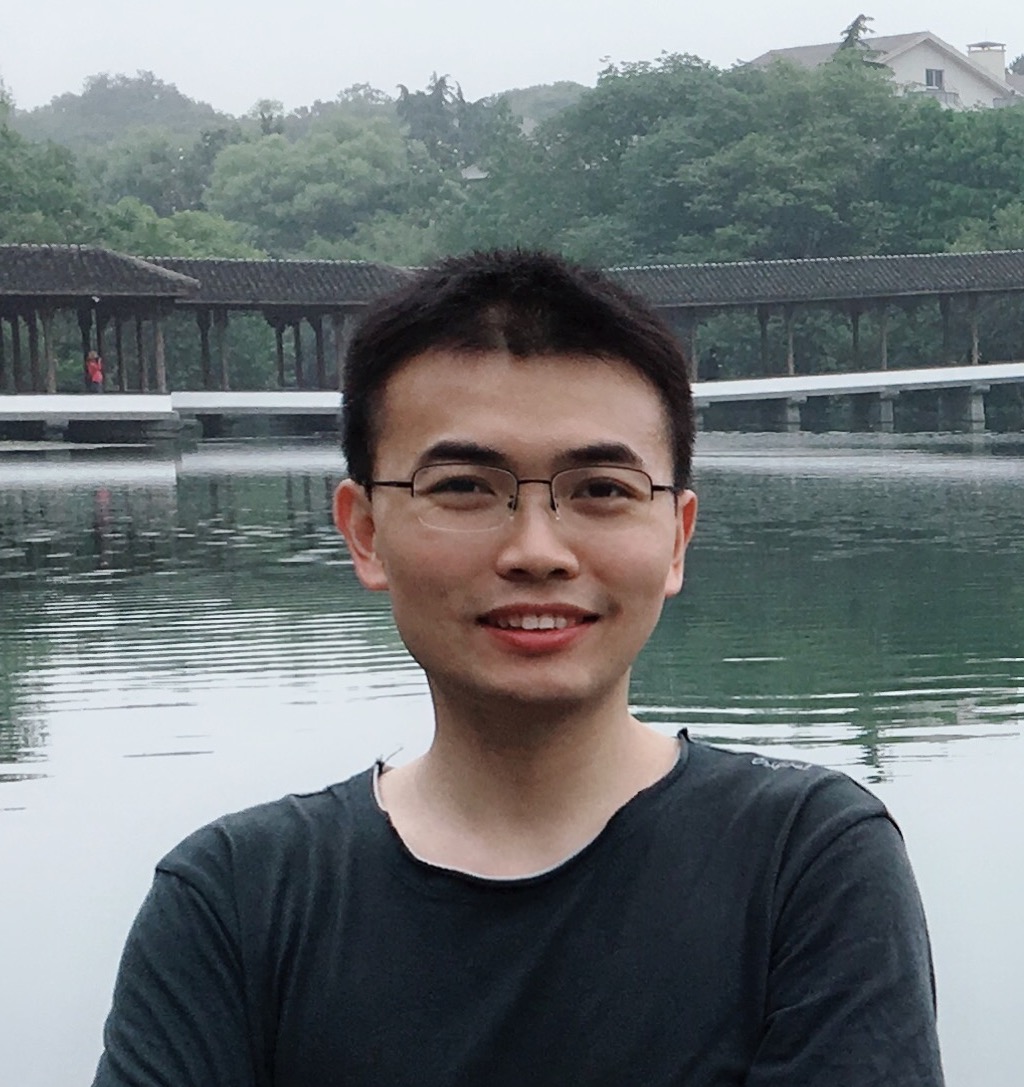}}]{Jiuniu Wang}
	received the B.Eng. degree in electrical engineering from Beijing Institute of Technology, Beijing, China, in 2016. He is currently a joint Ph.D. candidate in University of Chinese Academy of Sciences and City University of Hong Kong. His current research interests are in computer vision, natural language processing, and deep neural networks.
\end{IEEEbiography}

\begin{IEEEbiography}[{\includegraphics[width=1in,height=1.25in,clip,keepaspectratio]{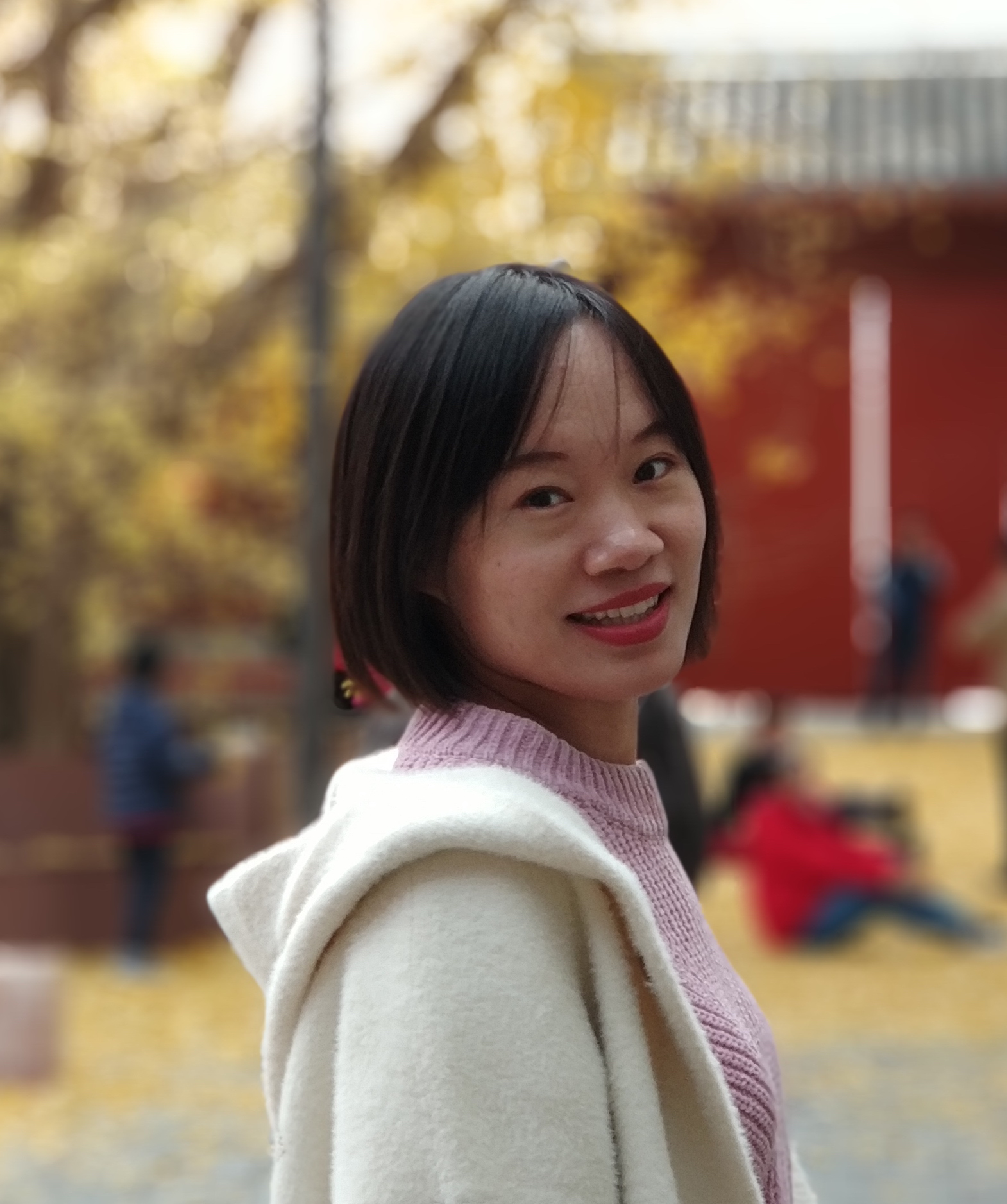}}]{Wenjia Xu} is a PhD student at the University of Chinese Academy of Sciences, China. She worked as a visiting PhD student at the Max-Planck Institute for Informatics in Saarbr\"ucken, Germany between 2019-2020. She received a B.Eng. degree with honors from Beijing Institute of Technology, China in 2016. Her research interest includes learning with limited supervision for computer vision tasks and explainable machine learning.
\end{IEEEbiography}

\begin{IEEEbiography}[{\includegraphics[width=1in,height=1.25in,clip,keepaspectratio]{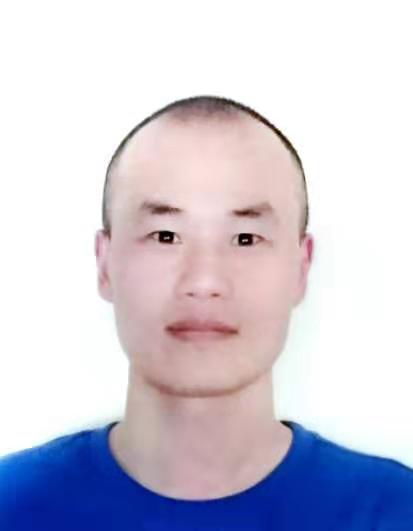}}]{Qinzhong Wang}
	received the B.Eng. and M.Eng. degrees in control science and engineering from Harbin Engineering University, Harbin, China, in 2013 and 2016, and Ph.D degree in 2021 from City University of Hong Kong, Hong Kong. Now he is a researcher at Baidu Research in Beijing, China. His research interests include computer vision, natural language processing, generative models and searching.
\end{IEEEbiography}

\begin{IEEEbiography}[{\includegraphics[width=1in,height=1.25in,clip,keepaspectratio]{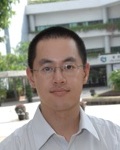}}]{Antoni B. Chan}
	received the B.S. and M.Eng. degrees in electrical engineering from Cornell University, Ithaca, NY, in 2000 and 2001, and the Ph.D. degree in electrical and computer engineering from the University of California, San Diego (UCSD), San Diego, in 2008. He is currently a Professor in the Department of Computer Science, City University of Hong Kong. His research interests include computer vision, machine learning, pattern recognition, and music analysis.
\end{IEEEbiography}

\appendices
\onecolumn
\begin{center}
	\LARGE
	On Distinctive Image Captioning via Comparing and Reweighting\\
	Supplementary material \\
\end{center}
\counterwithin{figure}{section}
\counterwithin{table}{section}

%

The organization of the supplementary material is as follows. 
We evaluate our Distinctiveness via Comparing and Reweighting~(DCR) model on the official online testing server in Appendix~\ref{sec:online}. More details of user study are introduced in Appendix~\ref{sec:sp_user_study}. In Appendix~\ref{section:sp_qualitative}, we present additional qualitative results.

\section{Online Evaluation}
\label{sec:online}
We evaluate our model on the official online test server in Table~\ref{table:online}. We compare our best model in Table~1 (\ie, Transformer+SCST+DCR) with other latest published captioning models. Note that we only use the single model on the online server, while other SOTA methods prefer to ensemble several models to obtain better results. Our model is designed to generate distinctive captions, while its accuracy is on par with other models aiming to match the quality metrics such as BLEU and CIDER.

\begin{table*}[!h]
	\begin{center}
		\caption{Leaderboard of the top ranking published state-of-the-art image captioning models on the online MS-COCO test server with 5 reference captions (c5) and 40 reference captions (c40).}	
		\label{table:online}
		\resizebox{450pt}{!}{
			\begin{tabular}{l|cc|cc|cc|cc|cc|cc|cc}
				\hline
				\multirow{2}{*}{Models}             & \multicolumn{2}{c|}{BLEU1} & \multicolumn{2}{c|}{BLEU2} & \multicolumn{2}{c|}{BLEU3} & \multicolumn{2}{c|}{BLEU4} & \multicolumn{2}{c|}{METEOR} & \multicolumn{2}{c|}{ROUGE-L} & \multicolumn{2}{c}{CIDEr-D} \\ \cline{2-15} 
				& c5           & c40         & c5           & c40         & c5           & c40         & c5           & c40         & c5           & c40          & c5            & c40          & c5           & c40          \\
				\hline
				SCST~\cite{rennie2017self}              & 78.1         & 93.7        & 61.9         & 86.0        & 47.0         & 75.9        & 35.2         & 64.5        & 27.0         & 35.5         & 56.3          & 70.7         & 114.7        & 116          \\
				LSTM-A~\cite{yao2017boosting}             & 78.7         & 93.7        & 62.7         & 86.7        & 47.6         & 76.5        & 35.6         & 65.2        & 27.0         & 35.4         & 56.4          & 70.5         & 116.0        & 118.0        \\
				UpDown~\cite{updown}             & 80.2         & 95.2        & 64.1         & 88.8        & 49.1         & 79.4        & 36.9         & 68.5        & 27.6         & 36.7         & 57.1          & 72.4         & 117.9        & 120.5        \\
				RFNet~\cite{jiang2018recurrent}              & 80.4        & 95.0        & 64.9         & 89.3        & 50.1         & 80.1        & 38.0         & 69.2        & 28.2         & 37.2         & 58.2          & 73.1         & 122.9        & 125.1        \\
				SGAE~\cite{yang2019auto}               & 81.0         & 95.3        & 65.6         & 89.5        & 50.7         & 80.4        & 38.5         & 69.7        & 28.2         & 37.2         & 58.6          & 73.6         & 123.8        & 126.5        \\
				AoANet~\cite{huang2019attention}             & 81.0         & 95.0        & 65.8         & 89.6        & 51.4         & 81.3        & \textbf{39.4}         & \textbf{71.2}        & \textbf{29.1}         & \textbf{38.5}         & 58.9          & \textbf{74.5}         & 126.9        & 129.6        \\
				GCN-LSTM+HIP~\cite{yao2019hierarchy}       & \textbf{81.6}         & \textbf{95.9}       & \textbf{66.2}         & \textbf{90.4}        & \textbf{51.5}         & \textbf{81.6}        & 39.3         & 71.0        & 28.8         & 38.1         & \textbf{59.0}          & 74.1         & \textbf{127.9}        & \textbf{130.2}        \\
				DCR~(Ours) & 80.2           &  94.6      & 64.8           &  88.6      & 50.2           &   79.8      & 38.3           &  69.4    & 28.6           &  37.8          & 58.3            &  73.4       & 123.8          & 126.0          \\ \hline
			\end{tabular}
		}
	\end{center}
\end{table*}

\section{User study interface}
\label{sec:sp_user_study}

\begin{figure}[!h]
	\centering
	\includegraphics[width=0.5\textwidth]{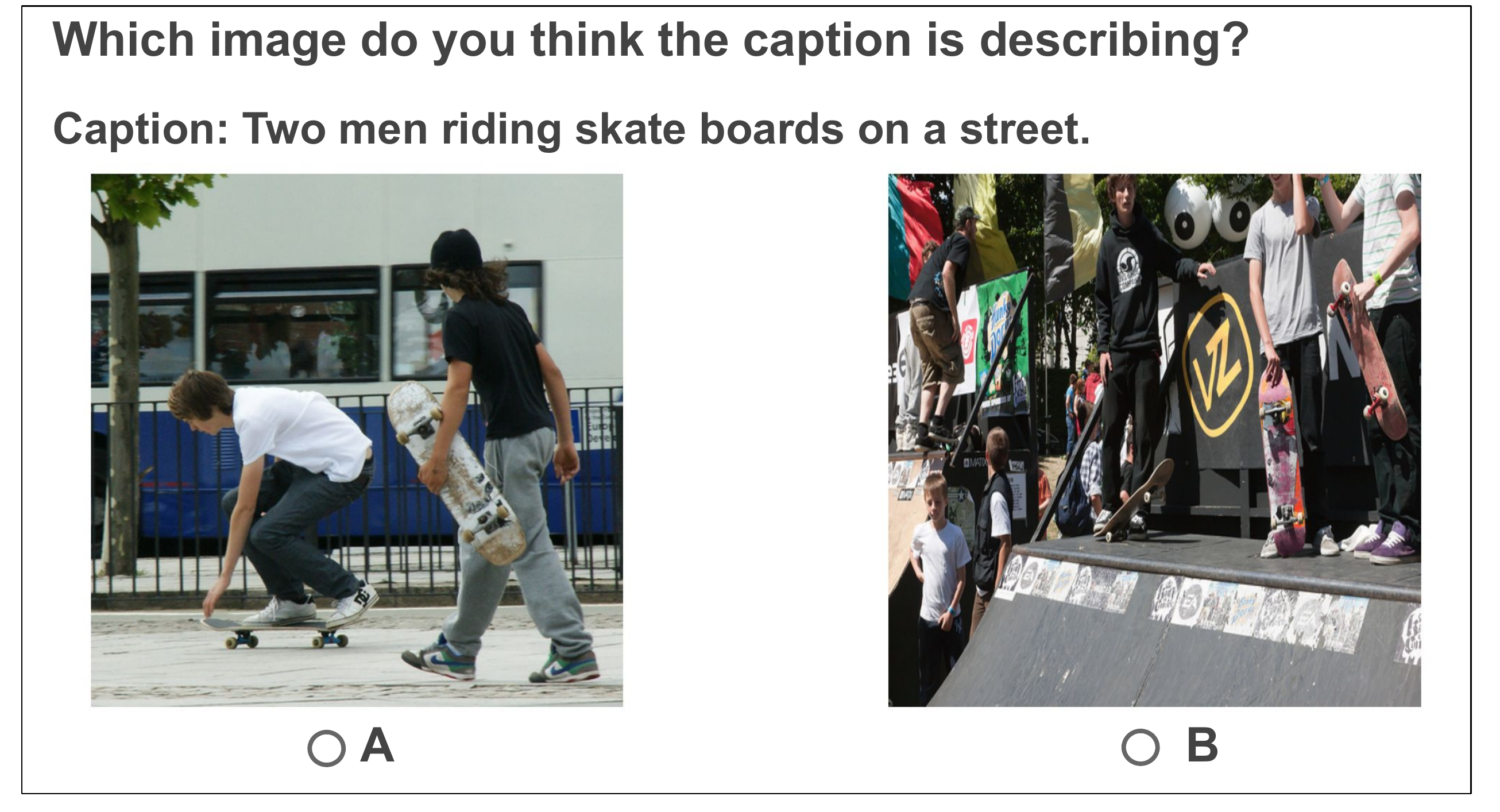}
	\caption{User study interface for image retrieval experiment.}
	\label{fig:Userstudy_interface1}
\end{figure}

\begin{figure}[!h]
	\centering
	\includegraphics[width=0.5\textwidth]{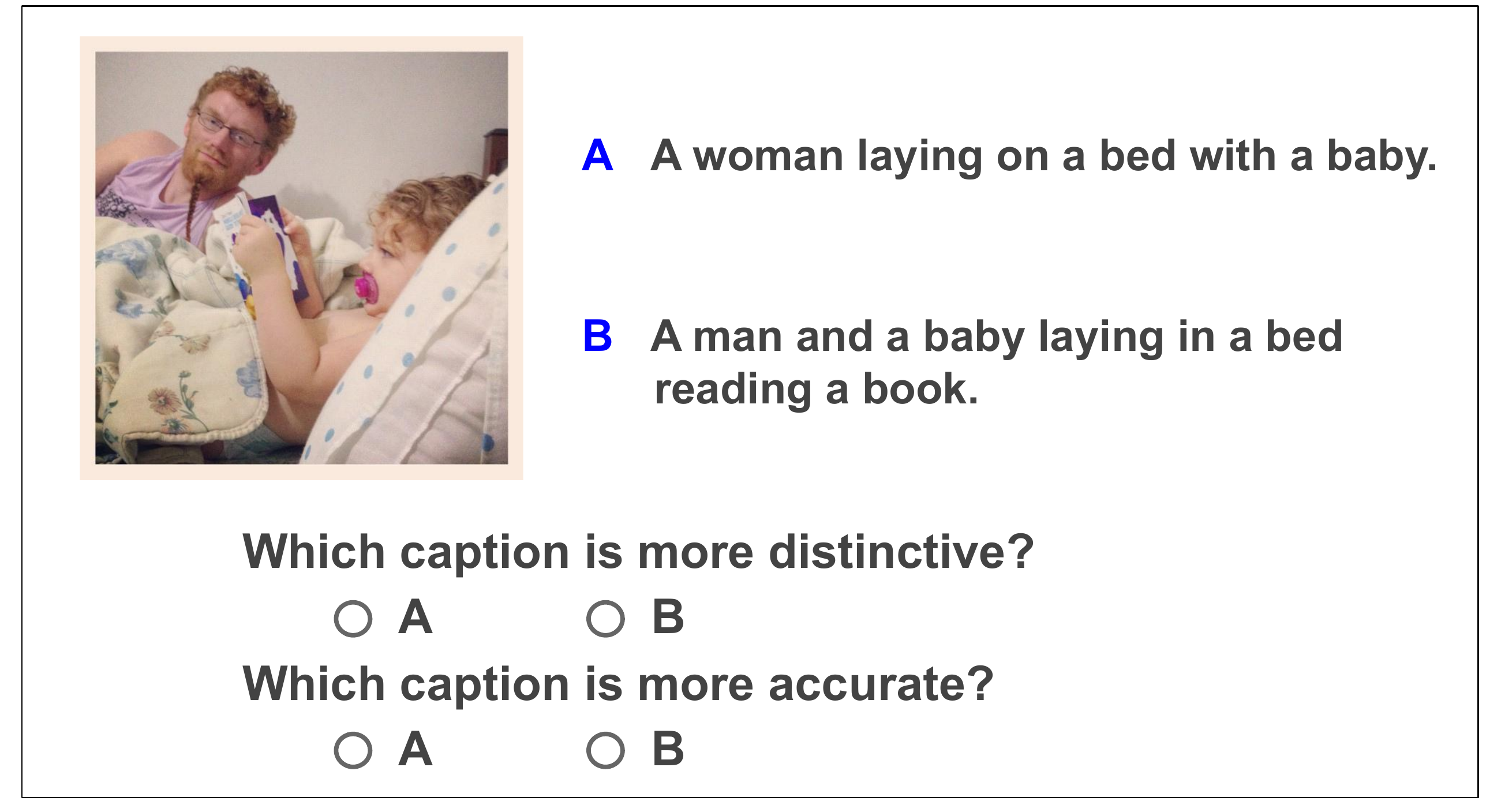}
	\caption{User study interface for selecting caption experiment.}
	\label{fig:Userstudy_interface2}
\end{figure}

To fairly evaluate the quality of the generated captions based on human perception, we conduct two user studies. The first user study can assess the distinctiveness of captions via image retrieval game. The task involves displaying the target image, a similar image with the same semantic meaning, and a generated caption describing the target image. Users are asked to choose the image that more closely matches the caption. The interface can be found in Figure~\ref{fig:Userstudy_interface1}. 
In the second user study, we ask participants to compare the two captions generated from a baseline model trained with and without DCR. In each trial, an image and two captions are displayed, and the user is asked to choose the better caption with respect to two criteria: distinctiveness and accuracy. The interface is shown in Figure~\ref{fig:Userstudy_interface2}.

\section{More Qualitative Results}
\label{section:sp_qualitative}

\begin{figure*}[!h]
	\centering
	\includegraphics[width=1\textwidth]{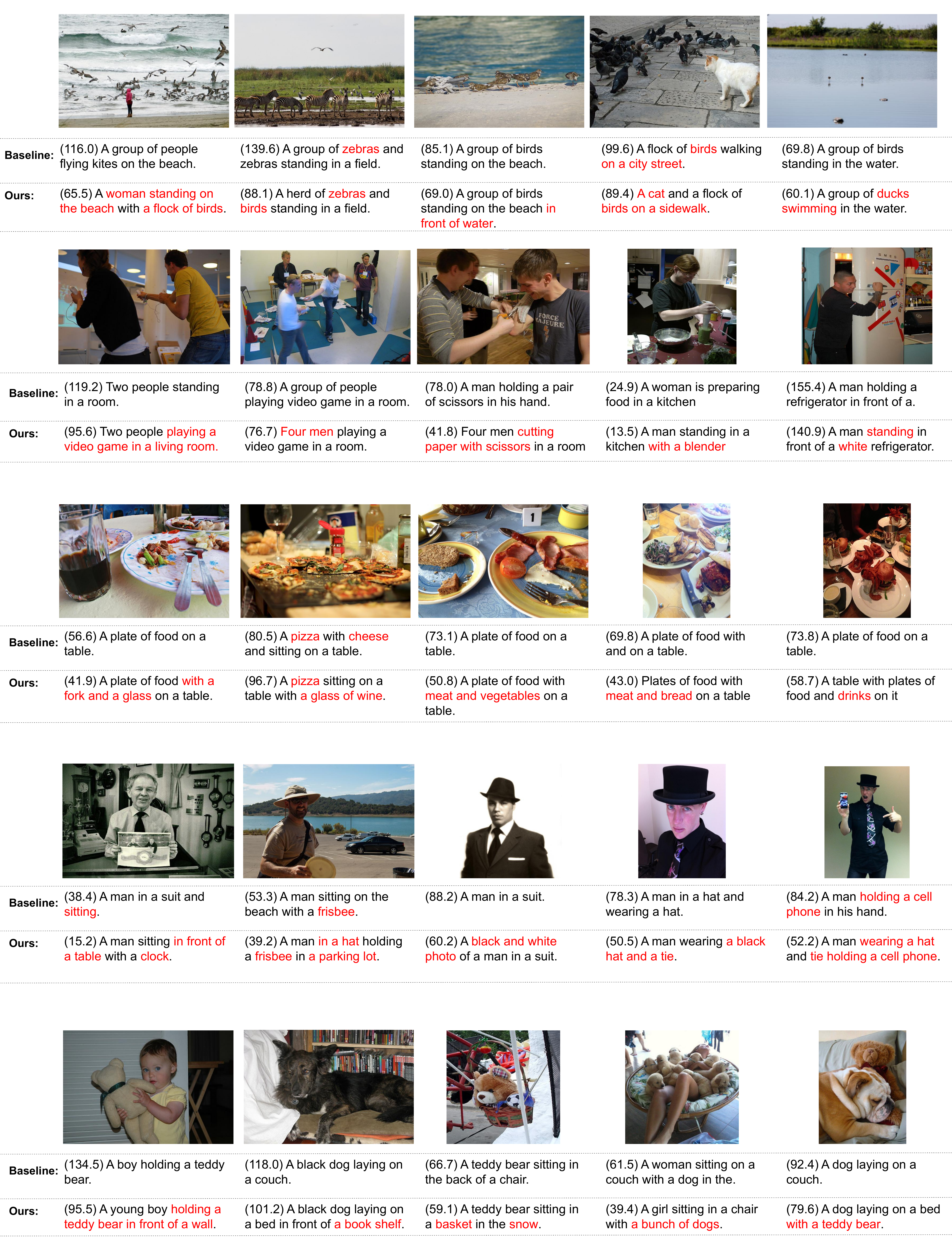}
	\caption{More qualitative results. Before each caption we report the CIDErBtw value representing the distinctiveness~(lower the better). We mark in red the distinctive words that are unique for the target image.}
	\label{fig:sp1}
\end{figure*}

\begin{figure*}[!h]
	\centering
	\includegraphics[width=1\textwidth]{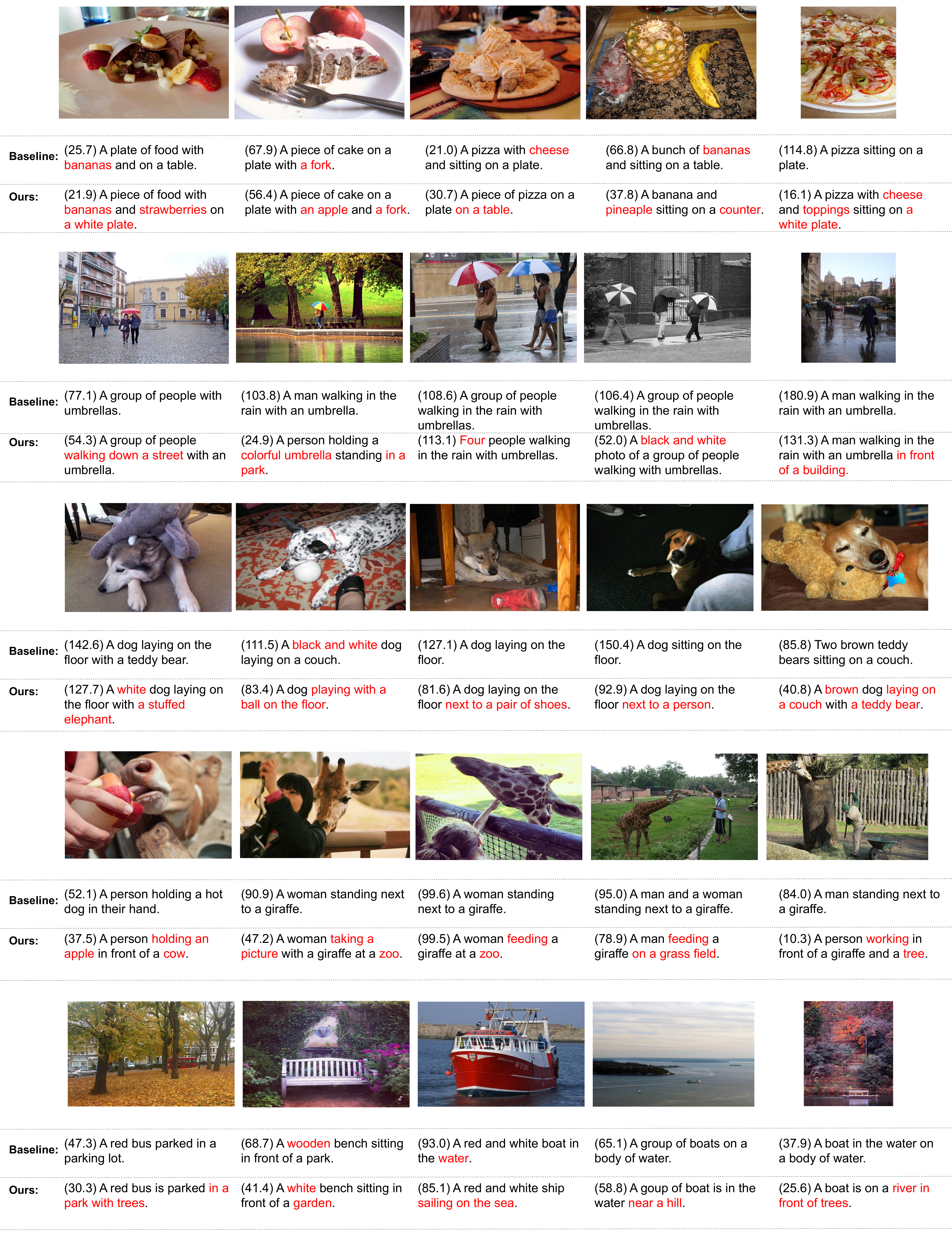}
	\caption{More qualitative results. Before each caption we report the CIDErBtw value representing the distinctiveness~(lower the better). We mark in red the distinctive words that are unique for the target image.}
	\label{fig:sp2}
\end{figure*}

We show more qualitative results in Figures~\ref{fig:sp1} and~\ref{fig:sp2}. In each row we display five images and their captions from one similar images set. ``Baseline'' denotes the captions generated from the Transformer model trained with SCST, and ``Ours'' denotes the captions from Transformer model trained with SCST and our DCR. Before each caption we report the CIDErBtw value representing the distinctiveness~(lower the better). We mark the unique words unique to the target image in red. Compared to the baseline model which produces general captions for all images, our model can generate captions with fruitful details that can distinguish each image from other similar images.


\end{document}